\documentclass[10pt,twocolumn,letterpaper]{article}

\usepackage{cvpr}              
\usepackage{booktabs}
\usepackage{multirow}
\usepackage{graphicx}

\usepackage{graphicx}
\usepackage{xcolor}
\usepackage{tikz}

\definecolor{attnblue}{RGB}{31,119,180}
\definecolor{topkgold}{RGB}{230,159,0}
\definecolor{coregreen}{RGB}{44,130,44}

\DeclareRobustCommand{\legendline}[1]{%
  \raisebox{0.2ex}{%
    \tikz{\draw[#1,line width=1.1pt] (0,0) -- (0.65,0);}%
  }%
}

\definecolor{cvprblue}{rgb}{0.21,0.49,0.74}
\usepackage[pagebackref,breaklinks,colorlinks,allcolors=cvprblue]{hyperref}

\def\confName{CVPR}
\def\confYear{2027}
\newcommand{\method}{\textsc{CoRE}}
\title{CoRE: Weakly Supervised Coarse-to-Fine Risk Evidence Learning \\in Driving Video}

\author{
Kaiser Hamid$^{1}$ \qquad
Can Cui$^{2}$ \qquad
Nade Liang$^{1}$\\
$^{1}$Texas Tech University
\qquad
$^{2}$Purdue University\\
\texttt{Project page: \url{https://kaiser-75.github.io/core/}}
}

\begin{document}
\maketitle
\begin{abstract}
Driver perceived risk evolves in dynamic environments and is supported 
by specific scene entities, yet supervision is typically limited to 
coarse video-level judgments. Learning \emph{when} supporting evidence 
emerges and \emph{which entities} support a risk predictor would 
require costly temporal- and entity-level annotations. We introduce 
\textbf{CoRE}, a weakly supervised coarse-to-fine framework that learns 
fine-grained prediction support from coarse video supervision. CoRE 
trains and freezes a video-level predictor. Structured interventions 
over candidate temporal regions or entity tracks measure how each 
candidate changes the coarse prediction, producing graded prediction-effect targets. These targets are distilled into a student that 
directly predicts temporal and entity support from the original video, 
without interventions at inference. We evaluate this learning principle
across three complementary settings. RISEE tests perceived-risk support 
from subjective clip-level judgments without temporal or entity-level 
risk annotations. DoTA provides independent temporal event annotations 
for evaluating weakly supervised traffic-anomaly localization. UCF-Crime tests whether the same coarse-to-fine mechanism extends to a
standard non-driving anomaly-detection benchmark. Across these
settings, CoRE learns informative fine-grained support from coarse
supervision, with strong temporal localization on DoTA and competitive
performance on UCF-Crime. These results establish coarse video 
predictions as supervision for recovering the fine-grained evidence 
supporting them, without corresponding fine-grained labels.
\end{abstract}    
\section{Introduction}
\label{sec:intro}

\begin{figure}[t]
    \centering
    \includegraphics[width=\columnwidth]{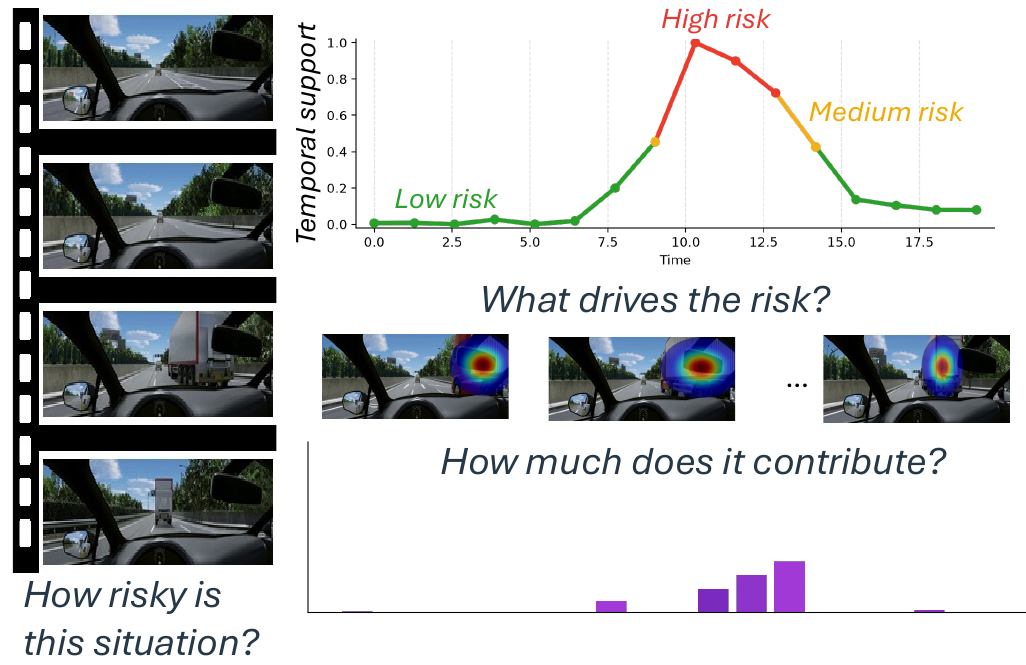}
    \caption{
    \textbf{CoRE} decomposes coarse video-level perceived risk into fine-grained
    support, revealing how risk evolves over time, what visual evidence drives
    the prediction, and how strongly individual objects contribute.
    }
    \label{fig:teaser}
\end{figure}

Modeling driver perceived risk can guide automated driving systems toward
decisions aligned with human perceived-risk judgments across diverse traffic
interactions~\cite{kolekar2020human}. A video-level score alone, however,
cannot reveal when supporting evidence emerges or which scene entities
support the prediction. Temporal support can inform warning timing, while
entity support can help prioritize relevant vehicles or pedestrians in the
surrounding traffic scene~\cite{Xu_2020_CVPR}. Yet egocentric driving videos
often provide only clip-level perceived-risk judgments or video-level event
labels, leaving this evidence unspecified. Because driving risk is
\emph{temporally evolving} and \emph{interaction-specific}, learning
\emph{when} and \emph{which entities} support a prediction from coarse
supervision offers a practical alternative to collecting costly temporal-
and entity-level annotations.

Weakly supervised video localization provides a natural starting point. In weakly supervised temporal action localization, CoLA improves ambiguous snippet representations \cite{zhang2021cola}, PivoTAL constructs localization-oriented supervision beyond classifier activation \cite{rizve2023pivotal}, and P-MIL learns directly over temporal proposals\cite{ren2023proposal}. Weakly supervised video anomaly detection similarly infers temporal anomaly scores from video-level labels through self-training \cite{feng2021mist}, feature-magnitude learning\cite{tian2021weakly}, and debiased multiple-instance learning \cite{lv2023unbiased}. These methods establish that coarse labels can support fine-grained prediction. However, their instance scores are still learned directly or indirectly from the bag-level objective and may emphasize the most discriminative snippets rather than the full evidence supporting the prediction. Accurate coarse prediction and reliable fine-grained support can therefore diverge.

This distinction is central to our problem. Video-level supervision constrains the coarse prediction but does not specify how support should be assigned across temporal regions and entities. Different support assignments can produce the same video-level response, leaving attention or multiple-instance scores without an explicit criterion for learning prediction support~\cite{jain2019attention}. Perturbation-based attribution measures such dependence more directly by modifying selected evidence and observing the resulting prediction change ~\cite{fong2017interpretable,fong2019understanding}. We use these measured effects to construct graded targets that provide an explicit criterion for assigning temporal and entity support. The student learns against these targets while retaining the original video-level objective.


We introduce \textbf{CoRE}, a weakly supervised coarse-to-fine framework based on \emph{prediction-effect distillation}. CoRE freezes a video-level predictor, measures prediction changes under structured interventions on temporal regions or entity tracks, and distills these effects into graded targets for a student that directly predicts temporal and entity support in a single forward pass. The learned support reflects prediction dependence rather than physical causality. We evaluate CoRE on RISEE~\cite{wu2025risee}, where only subjective clip-level perceived-risk judgments are available; DoTA~\cite{yao2022dota}, which provides independent temporal annotations in driving; and UCF-Crime~\cite{sultani2018real}, a standard non-driving anomaly benchmark, jointly testing fine-grained support, temporal localization, and generalization beyond driving.

Our contributions are:
\begin{itemize}
    \item We formulate \textbf{coarse-to-fine prediction-support
    learning for driving videos} and, to the best of our knowledge,
    provide the first formulation that learns both temporal and
    entity-level perceived-risk support from clip-level subjective
    judgments alone.

    \item We introduce \textbf{CoRE}, a prediction-effect distillation
    framework that converts the response of a frozen coarse predictor
    under structured candidate interventions into graded weak targets
    for direct temporal and entity support prediction.

    \item We validate CoRE on RISEE and DoTA for subjective risk
    support and independent temporal localization, respectively,
    and extend the evaluation to non-driving anomaly detection on 
    UCF-Crime.
\end{itemize}

\begin{figure*}[t]
\centering

\resizebox{\textwidth}{!}{%
    \includegraphics[height=5cm]{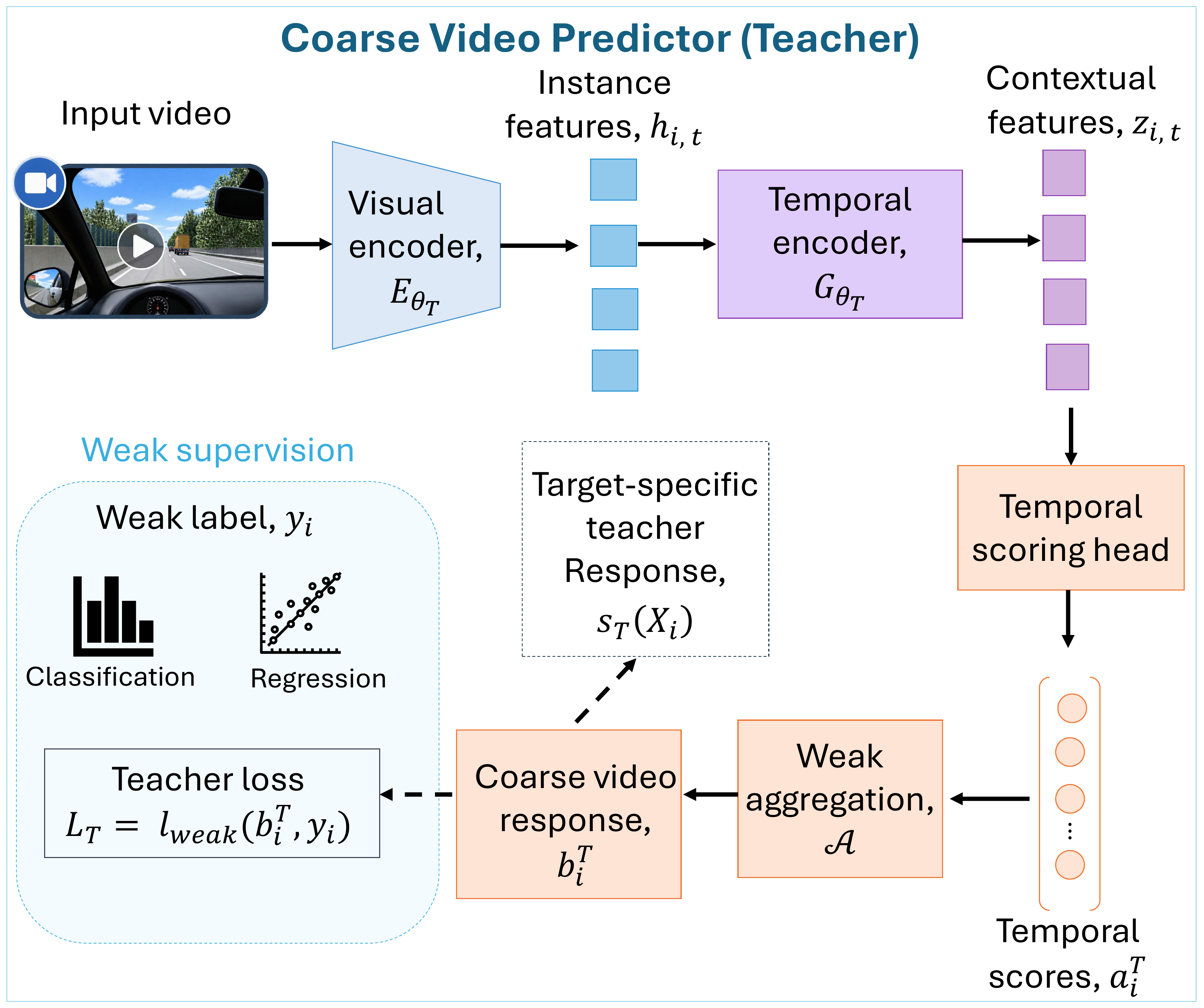}%
    \hspace{0.5mm}%
    \includegraphics[height=5cm]{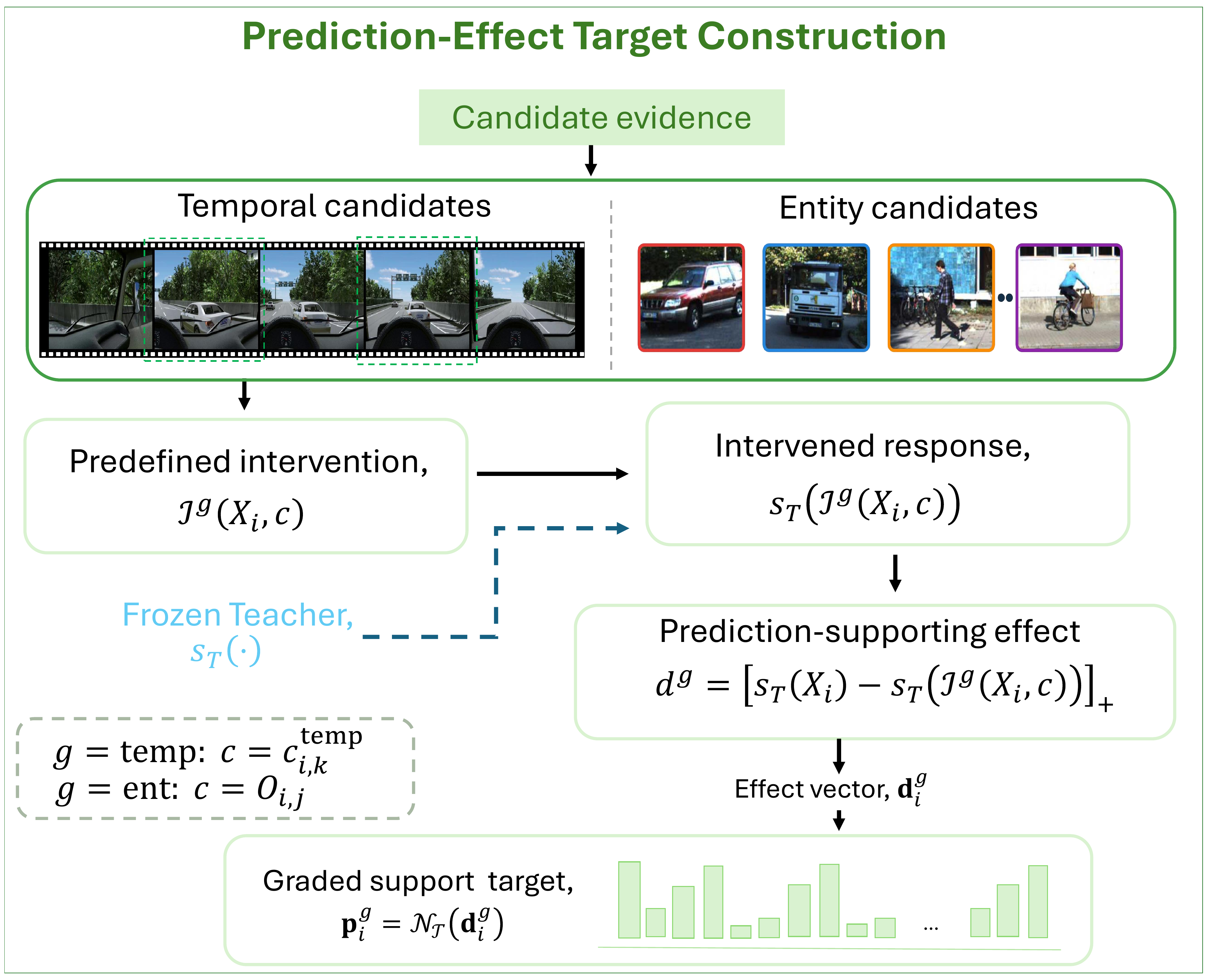}%
}

\vspace{2mm}

\resizebox{\textwidth}{!}{%
    \includegraphics[height=5cm]{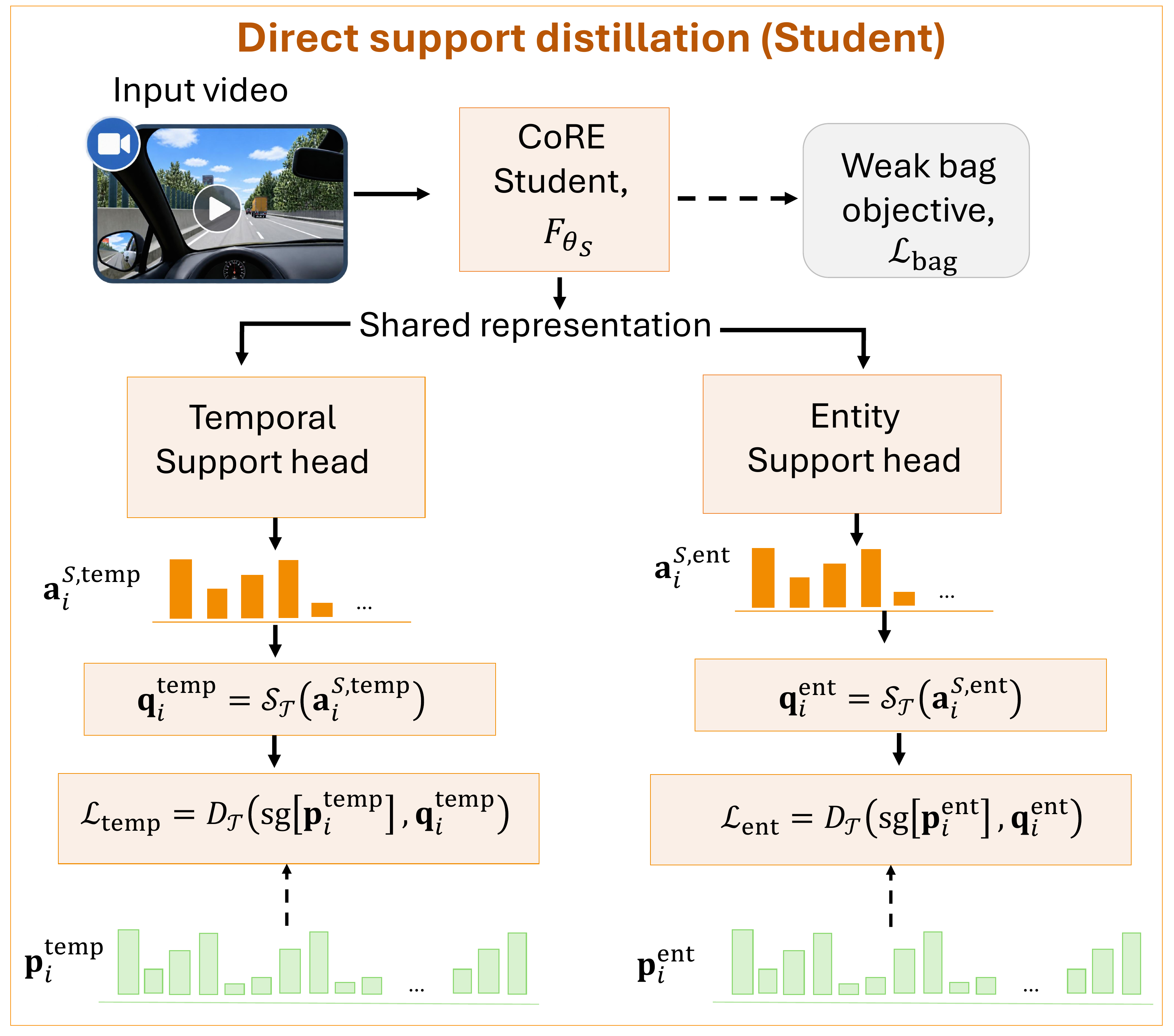}%
    \hspace{0.5mm}%
    \includegraphics[height=5cm]{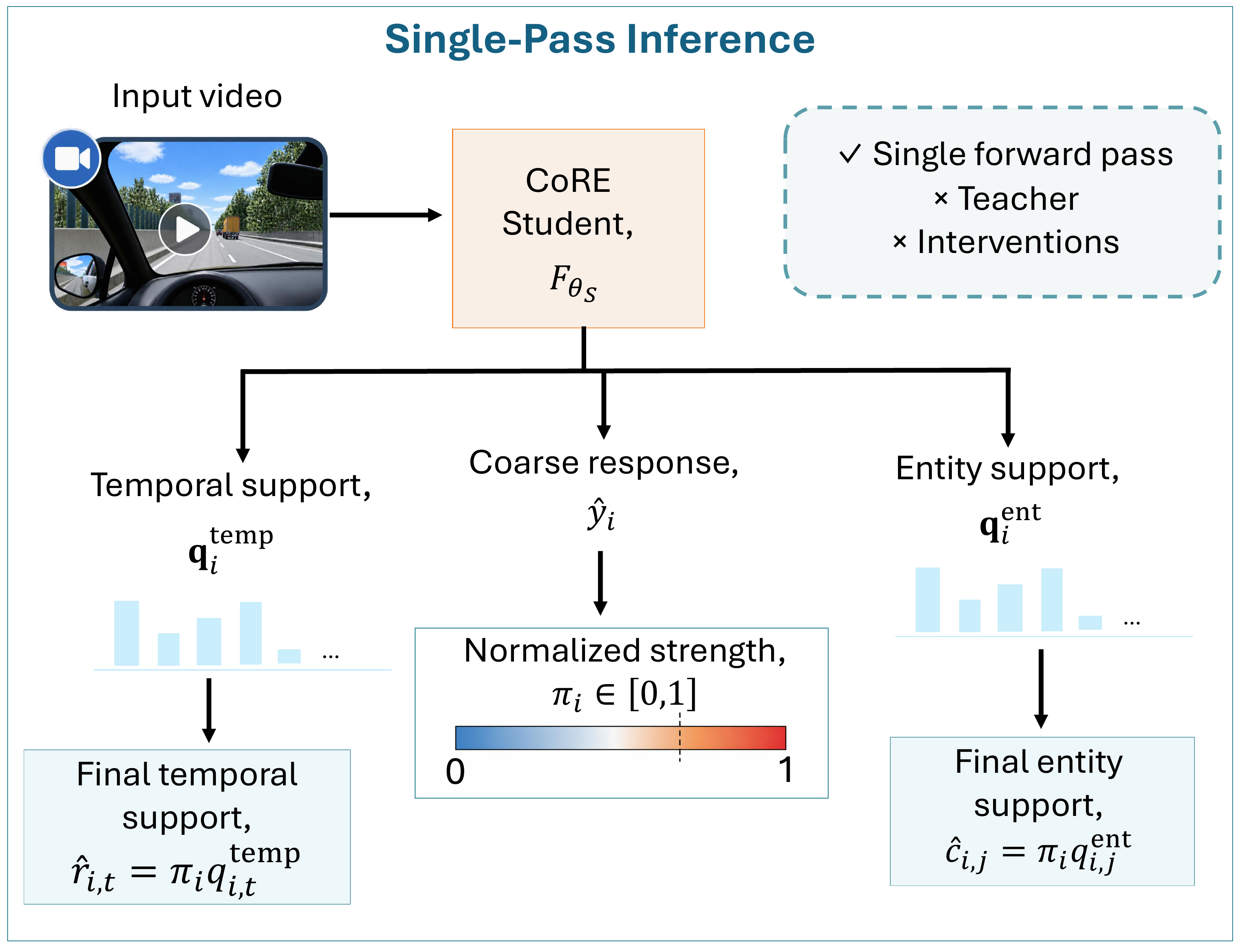}%
}

\caption{\textbf{Overview of CoRE.}
A coarse video predictor is trained from video-level supervision and
frozen. Structured interventions over temporal and entity candidates
produce prediction effects that are converted into graded support
targets and distilled into a student. At inference, the student
directly predicts the coarse response together with temporal and entity
support in a single forward pass.}
\label{fig:core_overview}
\end{figure*}

\section{Related Work}
\label{sec:related_work}

\textbf{Driving risk and accident understanding.}
Driving safety has been studied through accident anticipation, traffic-anomaly understanding, and the identification of influential scene entities. Early dash cam approaches model temporal context and traffic interactions to anticipate accidents before impact \cite{chan2016anticipating,zeng2017agent,suzuki2018anticipating}, while subsequent methods incorporate visual explanations and structured risk evolution \cite{bao2021drive,Zou_2026_CVPR}. Specifically, DoTA provides temporal, spatial, and categorical annotations for traffic anomalies \cite{yao2022dota}, whereas ROAD represents road events through structured agent--action--location labels \cite{singh2022road}.

A complementary line of work identifies traffic participants relevant to driving decisions. Prior methods identify influential scene entities by relating them to changes in driver behavior \cite{li2020make,li2023droid}. DRAMA and Rank2Tell provide important-object localization, ranking, and language supervision \cite{malla2023drama,sachdeva2024rank2tell}. MM-AU combines object-centric accident understanding with textual descriptions \cite{fang2024abductive}. More recently, RISEE introduces subjective perceived-risk judgments \cite{wu2025risee}, while RAID studies risk-object identification from driver-response supervision \cite{agarwal2026towards}. These methods often depend on task-specific supervision, including temporal event annotations, driver responses, or explicit object labels. Such supervision requires additional annotation effort and is unavailable in many video-level datasets. CoRE addresses this coarse-to-fine gap by recovering temporal and entity support from video-level supervision alone.

\textbf{Weakly supervised temporal localization.}
Weakly supervised temporal action localization recovers action intervals from video-level category labels. Early methods jointly learn video classification and temporal selection \cite{wang2017untrimmednets,nguyen2018weakly,paul2018w}, while later work addresses incomplete localization and foreground--background ambiguity through completeness modeling, iterative refinement, and stronger temporal representations \cite{liu2019completeness,pardo2021refineloc,luo2021action}. CoLA improves ambiguous snippets through contrastive learning \cite{zhang2021cola}, FTCL exploits fine-grained temporal structure \cite{gao2022fine}, and ASM-Loc models action-aware temporal segments \cite{he2022asm}. More direct coarse-to-fine approaches construct localization-oriented supervision. PivoTAL moves beyond localization by classification \cite{rizve2023pivotal}, P-MIL learns directly over temporal proposals \cite{ren2023proposal}, and PseudoFormer uses weak predictions to supervise a localization-oriented branch\cite{liu2025pseudoformer}. These methods demonstrate the value of supervision beyond raw classifier activations. CoRE differs in how it obtains this supervision. Candidate relevance is defined by the measured change in a trained coarse predictor under structured intervention. The same construction applies to both temporal regions and entity tracks.

\textbf{Weakly supervised video anomaly detection.}
Weakly supervised video anomaly detection learns fine-grained anomaly scores from video-level normal/anomalous labels. Sultani et al. introduced a multiple-instance ranking formulation in which normal and anomalous videos are treated as bags of temporal instances\cite{sultani2018real}. Subsequent methods improve instance discovery through pseudo-label self-training in MIST \cite{feng2021mist}, feature-magnitude learning in RTFM \cite{tian2021weakly}, normality-guided MIL \cite{10030221}, magnitude-contrastive representation learning in MGFN \cite{chen2022mgfnmagnitudecontrastiveglanceandfocusnetwork}, and debiased instance learning in UMIL \cite{lv2023unbiased}. PE-MIL incorporates language-derived priors into multiple-instance learning\cite{chen2024prompt}, while DAKD transfers aggregated knowledge across feature representations \cite{dalvi2025distilling}. Recent work has also examined how weakly supervised methods behave in egocentric driving-anomaly settings \cite{tiwari2024matters}. CoRE complements these approaches by obtaining temporal supervision from measured prediction effects. It measures the response of a frozen coarse predictor to structured candidate interventions and distills the resulting candidate-level prediction-effect distribution into a direct support model.

\textbf{Attribution, perturbation, and distillation.}
Attribution methods identify input evidence associated with a model prediction. Gradient-based approaches such as Integrated Gradients and Grad-CAM estimate relevance through local model sensitivity \cite{sundararajan2017axiomatic,selvaraju2017grad}, while attention weights may not reliably explain model decisions \cite{jain2019attention}. Perturbation-based approaches modify selected input content and measure the resulting prediction change \cite{fong2017interpretable,fong2019understanding}. These methods typically use perturbation only to analyze an existing model. CoRE uses measured prediction changes as training supervision. Candidate effects are converted into graded support targets and distilled into dedicated support predictors following the teacher--student principle \cite{hinton2015distilling}. Interventions are used only to construct training targets, while temporal and entity support are predicted directly at inference. The predicted support indicates which candidate evidence affects the coarse prediction. It does not establish physical causality.

\section{Method}
\label{sec:method}

\subsection{Problem Formulation}

We consider videos annotated only with a coarse video-level target.
Let
\begin{equation}
    X_i=\{x_{i,t}\}_{t=1}^{T_i}
\end{equation}
denote video $i$ with weak label $y_i$, where $x_{i,t}$ is a sampled
frame or snippet and $T_i$ is the number of samples. The label may
be continuous or categorical. Temporal boundaries and entity-level
relevance annotations are unavailable during training. Given $X_i$,
CoRE learns a video-level response $\hat y_i$, temporal support
$\mathbf q_i^{\mathrm{temp}}$, and, when entity candidates are
available, entity support $\mathbf q_i^{\mathrm{ent}}$.


CoRE follows a teacher--student procedure. A coarse predictor is trained
from video-level supervision and frozen. Structured interventions over
temporal or entity candidates measure their prediction effects, which
form graded targets for a separately initialized student. The student
predicts support directly from the original video, so interventions are
required only for target construction.

\subsection{Coarse Video Predictor}

A visual encoder extracts instance-level representations,
\begin{equation}
    h_{i,t}=E_{\theta_T}(x_{i,t}),
\end{equation}
where $\theta_T$ denotes the teacher parameters. A temporal encoder
then contextualizes these representations,
\begin{equation}
    z_{i,1:T_i}
    =
    G_{\theta_T}(h_{i,1:T_i}).
\end{equation}
A temporal scoring head produces instance logits
$\mathbf a_i^T$, which are summarized by a weak aggregation operator
$\mathcal A$ to obtain the coarse video response,
\begin{equation}
    b_i^T
    =
    \mathcal A(\mathbf a_i^T).
\end{equation}
The teacher is optimized with the available video-level supervision,
\begin{equation}
    \mathcal L_T
    =
    \ell_{\mathrm{weak}}(b_i^T,y_i),
\end{equation}
where $\ell_{\mathrm{weak}}$ is the corresponding regression or
classification objective.

The coarse predictor serves only as the source of prediction effects.
We denote by $s_T(X_i)$ the scalar teacher response associated with the
target being explained. For a continuous task, this is the normalized
scalar prediction; for a categorical task, it is the predicted score
or probability of the target class. Once coarse training is complete,
the teacher is frozen for all subsequent target construction.

\subsection{Prediction-Effect Target Construction}

Let
\begin{equation}
    \mathcal C_i^g
    =
    \{c_{i,1}^g,\ldots,c_{i,K_i^g}^g\},
    \qquad
    g\in\{\mathrm{temp},\mathrm{ent}\},
\end{equation}
denote a family of candidate evidence units. A temporal candidate
corresponds to a local temporal region, while an entity candidate
corresponds to a tracked scene element.

For candidate $c_{i,k}^g$, a predefined intervention
$\mathcal I^g(X_i,c_{i,k}^g)$ perturbs only that candidate while
preserving the remainder of the input. We measure its
prediction-supporting effect as
\begin{equation}
    d_{i,k}^g
    =
    \left[
        s_T(X_i)
        -
        s_T\left(
            \mathcal I^g(X_i,c_{i,k}^g)
        \right)
    \right]_+ ,
\label{eq:prediction_effect}
\end{equation}
where $[u]_+=\max(u,0)$. A larger value indicates stronger prediction
support under the specified intervention. We additionally evaluate robustness to alternative intervention operators.

Rather than retaining only the highest-effect candidate, CoRE uses the
full graded effect pattern. We convert the measured effects into a
support target through
\begin{equation}
    \mathbf p_i^g
    =
    \mathcal N_{\mathcal T}
    \left(
        \mathbf d_i^g
    \right),
\label{eq:support_target}
\end{equation}
where $\mathcal N_{\mathcal T}$ maps candidate effects to the support
representation required by the task. The mapping is determined by the support representation of the task.
This separates the shared CoRE mechanism from the final support
parameterization while retaining the same prediction-effect construction.

The resulting targets preserve relative effect strength instead of
collapsing the supervision to a single positive candidate. They are
computed using the frozen teacher, cached, and treated as fixed
supervision during student training.

\subsection{Temporal Support}

For temporal support, the candidate family consists of temporal
regions associated with the video sequence. Given the original,
unmodified video, the student temporal encoder produces contextual
features and a temporal support head predicts logits
$a_{i,t}^{S}$.

The logits are converted to temporal support scores by the
task-specific output map,
\begin{equation}
    \mathbf q_i^{\mathrm{temp}}
    =
    \mathcal S_{\mathcal T}
    \left(
        \mathbf a_i^{S,\mathrm{temp}}
    \right).
\end{equation}
The temporal head is trained to match the intervention-derived target,
\begin{equation}
    \mathcal L_{\mathrm{temp}}
    =
    D_{\mathcal T}
    \left(
        \operatorname{sg}
        [\mathbf p_i^{\mathrm{temp}}],
        \mathbf q_i^{\mathrm{temp}}
    \right),
\label{eq:temporal_distill}
\end{equation}
where $\operatorname{sg}[\cdot]$ denotes stop-gradient through the
teacher-derived target. The triplet
$(\mathcal N_{\mathcal T},\mathcal S_{\mathcal T},D_{\mathcal T})$
defines the support representation and corresponding matching loss
according to the task output; the prediction-effect construction
remains unchanged.

Importantly, the student never observes temporal annotations. Its
fine-grained supervision is derived entirely from the response of the
frozen coarse predictor to structured temporal interventions.

\subsection{Entity Support}

When entity candidates are available, the same construction is applied
at the entity level. Let
\begin{equation}
    \mathcal C_i^{\mathrm{ent}}
    =
    \{O_{i,1},\ldots,O_{i,J_i}\}
\end{equation}
denote the retained entity tracks. Each entity is represented using
the visual evidence associated with its track, and an entity-support
head predicts logits
$\mathbf a_i^{S,\mathrm{ent}}$.

The corresponding support scores are
\begin{equation}
    \mathbf q_i^{\mathrm{ent}}
    =
    \mathcal S_{\mathcal T}
    \left(
        \mathbf a_i^{S,\mathrm{ent}}
    \right),
\end{equation}
and are trained against the entity prediction-effect targets,
\begin{equation}
    \mathcal L_{\mathrm{ent}}
    =
    D_{\mathcal T}
    \left(
        \operatorname{sg}
        [\mathbf p_i^{\mathrm{ent}}],
        \mathbf q_i^{\mathrm{ent}}
    \right).
\end{equation}

Temporal regions and entity tracks therefore use the same learning
rule: define candidate evidence, measure how intervening on each
candidate changes the coarse prediction, convert those effects into
graded support targets, and distill them into a direct predictor.
The entity branch is permutation-equivariant through shared track
encoding and scoring without candidate-index embeddings, while
symmetric masked normalization yields relative support over the
retained candidates and context token.

\subsection{Support Distillation and Inference}

The student retains the original video-level objective while learning
fine-grained support. We write the complete training objective as
\begin{equation}
    \mathcal L_{\mathrm{CoRE}}
    =
    \lambda_{\mathrm{eff}}
    \mathcal L_{\mathrm{effect}}
    +
    \lambda_{\mathrm{bag}}
    \mathcal L_{\mathrm{bag}}
    +
    \lambda_{\mathrm{reg}}
    \mathcal L_{\mathrm{reg}},
\label{eq:core_objective}
\end{equation}
with
\begin{equation}
    \mathcal L_{\mathrm{effect}}
    =
    \mathcal L_{\mathrm{temp}}
    +
    \lambda_e\mathcal L_{\mathrm{ent}},
\end{equation}
where $\lambda_e=0$ when entity candidates are unavailable.
$\mathcal L_{\mathrm{bag}}$ preserves the original weak video-level
prediction objective, while $\mathcal L_{\mathrm{reg}}$ contains the
standard localization regularization used during student training.
Exact task configurations, regularization terms, and loss weights are provided in
the supplementary material.

At inference, the teacher and all intervention operations are removed.
The student processes the original video once and directly predicts
the coarse response together with temporal and, when available, entity
support.

Let $\pi_i\in[0,1]$ denote the normalized strength of the coarse
prediction. The final temporal support is
\begin{equation}
    \hat r_{i,t}
    =
    \pi_i q_{i,t}^{\mathrm{temp}},
\end{equation}
and the entity support is
\begin{equation}
    \hat c_{i,j}
    =
    \pi_i q_{i,j}^{\mathrm{ent}}.
\end{equation}
Thus, the fine-grained scores describe where the prediction is
supported, while the coarse response determines the overall prediction
strength.

CoRE therefore separates coarse prediction from support learning. The
coarse objective and support output follow the task definition, while
the central procedure---structured intervention, prediction-effect
measurement, graded target construction, and direct student
distillation---is shared throughout.
\section{Experiments}
\label{sec:experiments}
We evaluate \method{} in three complementary weakly supervised video
settings. RISEE~\cite{wu2025risee} studies perceived-risk support from
clip-level subjective ratings, DoTA~\cite{yao2022dota} provides
independent temporal annotations in driving, and
UCF-Crime~\cite{sultani2018real} tests extension to a standard
non-driving anomaly benchmark. Fine-grained annotations are never
used for training.

\subsection{Experimental Setup}
\label{sec:experimental_setup}

\noindent\textbf{Datasets and protocols.}
RISEE contains $179$ egocentric driving scenarios with aggregated
human perceived-risk ratings. We use scenario-level five-fold
evaluation and preserve identical folds across all compared methods.
The clip-level human rating is the only supervision used for
perceived-risk learning. Since RISEE provides neither temporal risk
intervals nor risk-entity labels, the primary fidelity evaluation therefore 
uses held-out candidate interventions. Supplement~E compares 
out-of-fold CoRE temporal support with expert-consensus temporal intervals on a
$30$-video subset.

For DoTA, we follow the weakly supervised reorganization of
Tiwari et al.~\cite{tiwari2024matters}. The training pool combines
anomalous DoTA videos with normal driving videos from $D^2$-City. We
use $2{,}420$ anomalous and $3{,}234$ normal videos for training,
reserve $269$ anomalous and $358$ normal videos for validation, and
retain the released $1{,}140$-video DoTA test split, which contains
only anomaly-containing videos and therefore supports temporal
localization but not binary clip-level classification. 

For UCF-Crime~\cite{sultani2018real}, we use the standard split of
$1{,}610$ training and $290$ test videos. Training uses only
video-level normal/abnormal labels, and frame-level intervals are used
only for test evaluation. Following the MGFN data
protocol~\cite{chen2022mgfnmagnitudecontrastiveglanceandfocusnetwork},
CoRE uses fixed, pre-extracted ten-crop, $2048$-dimensional I3D RGB
features arranged into $32$-segment sequences. We also run
RTFM~\cite{tian2021weakly} and MGFN using their official repositories
and standard UCF-Crime configurations.

\noindent\textbf{Implementation details.}
Across settings, CoRE follows the procedure in
Sec.~\ref{sec:method}: a coarse predictor is learned from video-level
supervision, the frozen predictor generates candidate-level
prediction-effect targets, and a separately initialized student learns
to predict fine-grained support directly from the original input. The
coarse prediction head and support output follow the corresponding
task definition, while prediction-effect construction remains shared.
For RISEE, we use a frozen, pretrained ResNet-50 encoder and activate
the entity-support branch. For DoTA, we
evaluate separate ResNet-50 and CLIP ViT-B/32 feature banks. For
UCF-Crime, we use fixed, pre-extracted I3D RGB features. Task configurations,
training settings, and loss weights are provided
in the supplementary material.

\noindent\textbf{Baselines and metrics.}
On RISEE, we compare clip-level perceived-risk prediction with
frame-average regression, Attention MIL~\cite{ilse2018attention},
Soft top-$k$ MIL~\cite{sultani2018real}, temporal convolution, a
temporal Transformer, and Video Swin
regression~\cite{liu2022video}. We report MAE and RMSE for prediction
error, together with Spearman $\rho$ and pairwise ranking accuracy
(PairAcc) for agreement with the ordering of the clip-level ratings.
For temporal prediction support, we compare against Uniform weighting,
Attention MIL~\cite{ilse2018attention}, and Soft top-$k$
MIL~\cite{sultani2018real}. We report selected prediction drop
(Sel.\ Drop), gain over a size-matched random intervention
(Gain/Rand.), correlation with measured candidate effects
(Effect $\rho$), the selected window's percentile among measured
effects (Top Perc.), and the fraction of selections producing a
positive prediction drop (Pos.\ Rate). Entity-support evaluation
additionally uses NDCG@3 and Regret@1.

On DoTA, we reproduce or adapt Deep MIL~\cite{sultani2018real},
RTFM~\cite{tian2021weakly},
MGFN~\cite{chen2022mgfnmagnitudecontrastiveglanceandfocusnetwork},
UR-DMU~\cite{zhou2023dual}, OE-CTST~\cite{majhi2024oe},
PE-MIL~\cite{chen2024prompt}, and TPWNG~\cite{yang2024text} under the
same WS-DoTA split and evaluation pipeline. Within each feature block,
all methods use the same visual features, WS-DoTA split, and evaluation pipeline.
We report frame AUC, frame AP,
macro AUC, event-level F1 at temporal-IoU thresholds $0.3$ and $0.5$,
and best temporal IoU.

On UCF-Crime, we compare with MIL-Rank~\cite{sultani2018real},
MIST~\cite{feng2021mist}, RTFM~\cite{tian2021weakly},
NG-MIL~\cite{10030221},
MGFN~\cite{chen2022mgfnmagnitudecontrastiveglanceandfocusnetwork},
DSS~\cite{xu2025discriminative}, PE-MIL~\cite{chen2024prompt},
$\pi$-VAD~\cite{majhi2025just},
LEC-VAD~\cite{wang2025learning}, and
LAS-VAD~\cite{wang2026weakly}. We separate methods trained only with
RGB features and video-level normal/anomalous labels from methods that
also use text or auxiliary visual cues during training. Following the
standard protocol, we report frame-level AUC.

\subsection{Main Results}
\label{sec:main_results}

\noindent\textbf{Perceived-risk prediction and support on RISEE.}
Table~\ref{tab:risee_score} first evaluates the directly supervised
clip-level task. CoRE obtains the lowest MAE and highest Spearman
correlation while remaining competitive on RMSE and PairAcc,
showing that fine-grained support learning does not compromise the
underlying perceived-risk prediction. The stronger distinction appears in Table~\ref{tab:risee_temporal}.
Although Attention MIL and Soft top-$k$ MIL provide useful video-level
instance scores, their selected regions produce small prediction drops
and negative correlation with measured window effects. CoRE instead
achieves a $0.320$ selected drop, $0.245$ gain over random, and
$0.542$ effect correlation. Thus, instance scores sufficient for
forming a coarse prediction need not identify the evidence on which
that prediction depends.

\begin{table}[!t]
\centering
\caption{\textbf{Clip-level perceived-risk prediction on RISEE.}
Scenario-level five-fold evaluation using only human clip-level
ratings. Best per metric in \textbf{bold}, second best
\underline{underlined}.}
\label{tab:risee_score}
\scriptsize
\setlength{\tabcolsep}{3.5pt}
\resizebox{\columnwidth}{!}{%
\begin{tabular}{lcccc}
\toprule
Method
& MAE $\downarrow$
& RMSE $\downarrow$
& Spearman $\rho$ $\uparrow$
& PairAcc $\uparrow$ \\
\midrule
Frame-average regression
& 0.640 & 0.788 & 0.581 & 0.721 \\
Attention MIL~\cite{ilse2018attention}
& \underline{0.577} & \textbf{0.709}
& \underline{0.653} & \textbf{0.750} \\
Soft top-$k$ MIL~\cite{sultani2018real}
& 0.683 & 0.815 & 0.575 & 0.713 \\
Temporal convolution
& 0.649 & 0.808 & 0.572 & 0.710 \\
Temporal Transformer
& 0.659 & 0.801 & 0.552 & 0.703 \\
Video Swin regression~\cite{liu2022video}
& 0.592 & \underline{0.718} & 0.641 & 0.742 \\
\midrule
\textbf{\method{} (ours)}
& \textbf{0.575} & 0.724 & \textbf{0.655}
& \underline{0.748} \\
\bottomrule
\end{tabular}%
}
\end{table}

\begin{table}[!t]
\centering
\caption{\textbf{Temporal prediction support on RISEE.}
Held-out interventions measure prediction dependence; Effect $\rho$
measures agreement with candidate effects, while Top Perc.\ and
Pos.\ Rate summarize selection quality.}
\label{tab:risee_temporal}
\scriptsize
\setlength{\tabcolsep}{3.0pt}
\resizebox{\columnwidth}{!}{%
\begin{tabular}{lccccc}
\toprule
Method
& Sel. Drop $\uparrow$
& Gain/Rand. $\uparrow$
& Effect $\rho$ $\uparrow$
& Top Perc. $\uparrow$
& Pos. Rate $\uparrow$ \\
\midrule
Uniform weighting
& 0.045 & -0.049 & 0.000 & 0.538 & 0.726 \\
Attention MIL~\cite{ilse2018attention}
& 0.032 & -0.050 & -0.167 & 0.482 & 0.620 \\
Soft top-$k$ MIL~\cite{sultani2018real}
& 0.050 & -0.038 & -0.214 & 0.474 & 0.592 \\
\midrule
\textbf{\method{} (ours)}
& \textbf{0.320}
& \textbf{0.245}
& \textbf{0.542}
& \textbf{0.888}
& \textbf{0.989} \\
\bottomrule
\end{tabular}%
}
\end{table}

\begin{table*}[t]
\centering
\caption{\textbf{Weakly supervised temporal anomaly localization on
DoTA.} All results use the same WS-DoTA split and evaluation pipeline.
Methods within each block share the same visual feature bank.
$^{\dagger}$ denotes a protocol-matched adaptation based on released
code; $^{\ddagger}$ denotes our reimplementation from the
paper/supplement when executable official code was unavailable;
$^{\S}$ denotes our controlled baseline. Best in \textbf{bold};
second best \underline{underlined}.}
\label{tab:dota_main_comparison}
\scriptsize
\setlength{\tabcolsep}{3.8pt}
\begin{tabular}{lllcccccc}
\toprule
Features
& Method
& Venue
& Frame AUC $\uparrow$
& Frame AP $\uparrow$
& Macro AUC $\uparrow$
& F1@0.3 $\uparrow$
& F1@0.5 $\uparrow$
& Best tIoU $\uparrow$ \\
\midrule

\multirow{9}{*}{ResNet-50}
& Top-$k$ MIL$^{\S}$ &
& 0.377 & 0.251 & 0.381 & 0.499 & 0.147 & 0.327 \\

& Deep MIL$^{\dagger}$~\cite{sultani2018real} & CVPR'18
& 0.520 & 0.335 & 0.526 & 0.462 & 0.150 & 0.314 \\

& RTFM$^{\dagger}$~\cite{tian2021weakly} & ICCV'21
& 0.565 & 0.349 & 0.569 & 0.543 & 0.191 & 0.352 \\

& MGFN$^{\dagger}$~
\cite{chen2022mgfnmagnitudecontrastiveglanceandfocusnetwork}
& AAAI'23
& \underline{0.642} & 0.407 & \underline{0.635}
& 0.493 & 0.149 & 0.322 \\

& UR-DMU$^{\dagger}$~\cite{zhou2023dual} & AAAI'23
& 0.574 & 0.398 & 0.582 & 0.488 & 0.143 & 0.316 \\

& OE-CTST$^{\ddagger}$~\cite{majhi2024oe} & WACV'24
& 0.387 & 0.257 & 0.396 & 0.349 & 0.139 & 0.361 \\

& PE-MIL$^{\ddagger}$~\cite{chen2024prompt} & CVPR'24
& 0.618 & \underline{0.410} & 0.610
& \underline{0.562} & 0.184 & 0.345 \\

& TPWNG$^{\ddagger}$~\cite{yang2024text} & CVPR'24
& 0.578 & 0.367 & 0.566 & 0.507
& \underline{0.237} & \underline{0.370} \\

\cmidrule(lr){2-9}
& \textbf{\method{} (ours)} &
& \textbf{0.735} & \textbf{0.484} & \textbf{0.744}
& \textbf{0.747} & \textbf{0.374} & \textbf{0.440} \\

\midrule

\multirow{9}{*}{CLIP ViT-B/32}
& Top-$k$ MIL$^{\S}$ &
& 0.492 & 0.310 & 0.487 & 0.502 & 0.142 & 0.325 \\

& Deep MIL$^{\dagger}$~\cite{sultani2018real} & CVPR'18
& 0.507 & 0.330 & 0.507 & 0.497 & 0.152 & 0.333 \\

& RTFM$^{\dagger}$~\cite{tian2021weakly} & ICCV'21
& 0.610 & 0.395 & 0.614 & 0.586 & 0.240 & 0.381 \\

& MGFN$^{\dagger}$~
\cite{chen2022mgfnmagnitudecontrastiveglanceandfocusnetwork}
& AAAI'23
& \underline{0.671} & \underline{0.429}
& \underline{0.674} & \underline{0.647}
& \underline{0.258} & \underline{0.391} \\

& UR-DMU$^{\dagger}$~\cite{zhou2023dual} & AAAI'23
& 0.502 & 0.328 & 0.501 & 0.493 & 0.144 & 0.323 \\

& OE-CTST$^{\ddagger}$~\cite{majhi2024oe} & WACV'24
& 0.608 & 0.396 & 0.607 & 0.597 & 0.197 & 0.356 \\

& PE-MIL$^{\ddagger}$~\cite{chen2024prompt} & CVPR'24
& 0.471 & 0.288 & 0.461 & 0.492 & 0.151 & 0.312 \\

& TPWNG$^{\ddagger}$~\cite{yang2024text} & CVPR'24
& 0.629 & 0.414 & 0.620 & 0.524 & 0.209 & 0.338 \\

\cmidrule(lr){2-9}
& \textbf{\method{} (ours)} &
& \textbf{0.744} & \textbf{0.514} & \textbf{0.743}
& \textbf{0.719} & \textbf{0.364} & \textbf{0.429} \\
\bottomrule
\end{tabular}
\end{table*}
\begin{figure*}[!tb]
    \centering

    \includegraphics[width=0.44\textwidth]{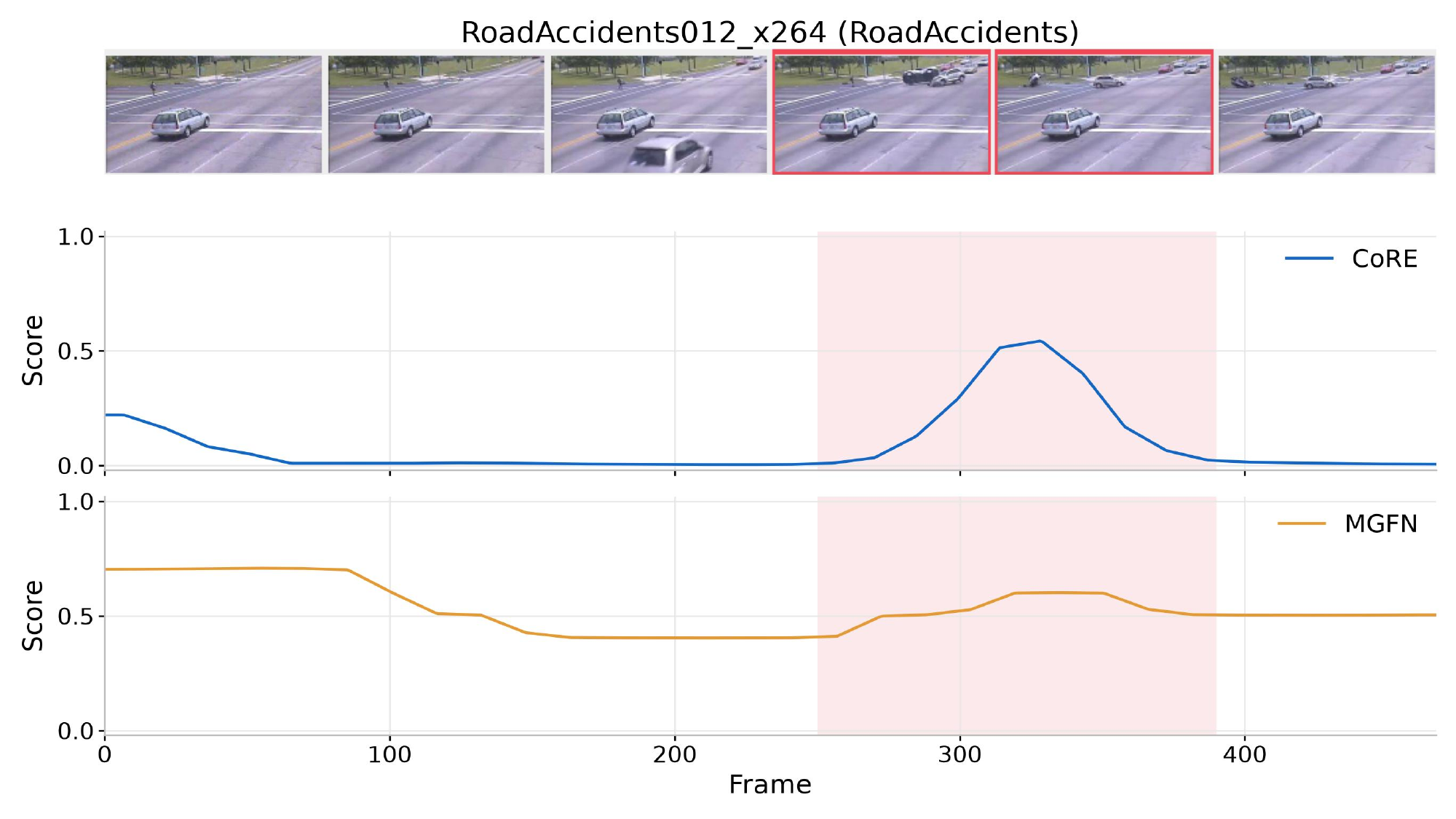}
    \hfill
    \includegraphics[width=0.44\textwidth]{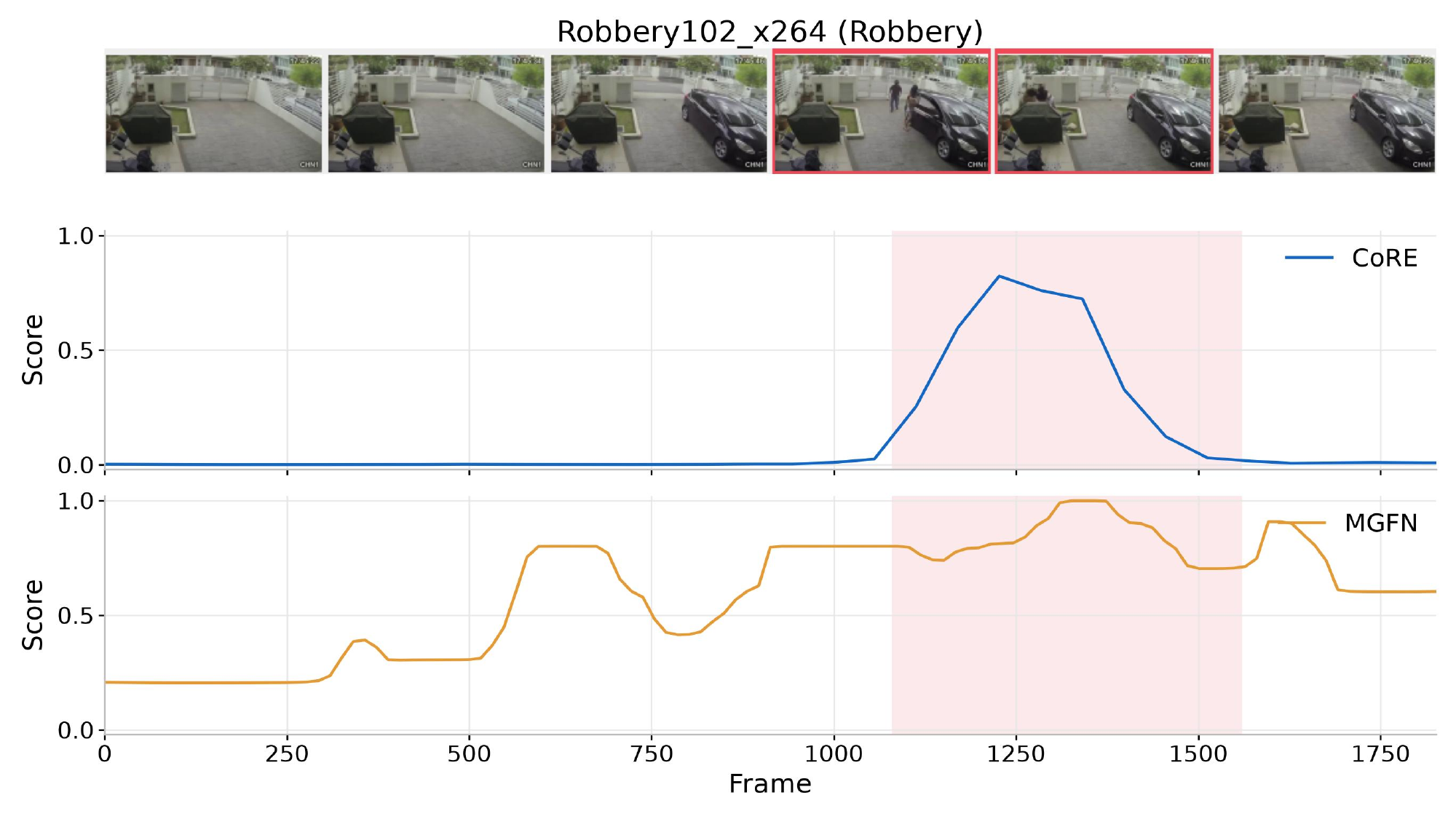}

    \vspace{0.25em}

    \includegraphics[width=0.44\textwidth]{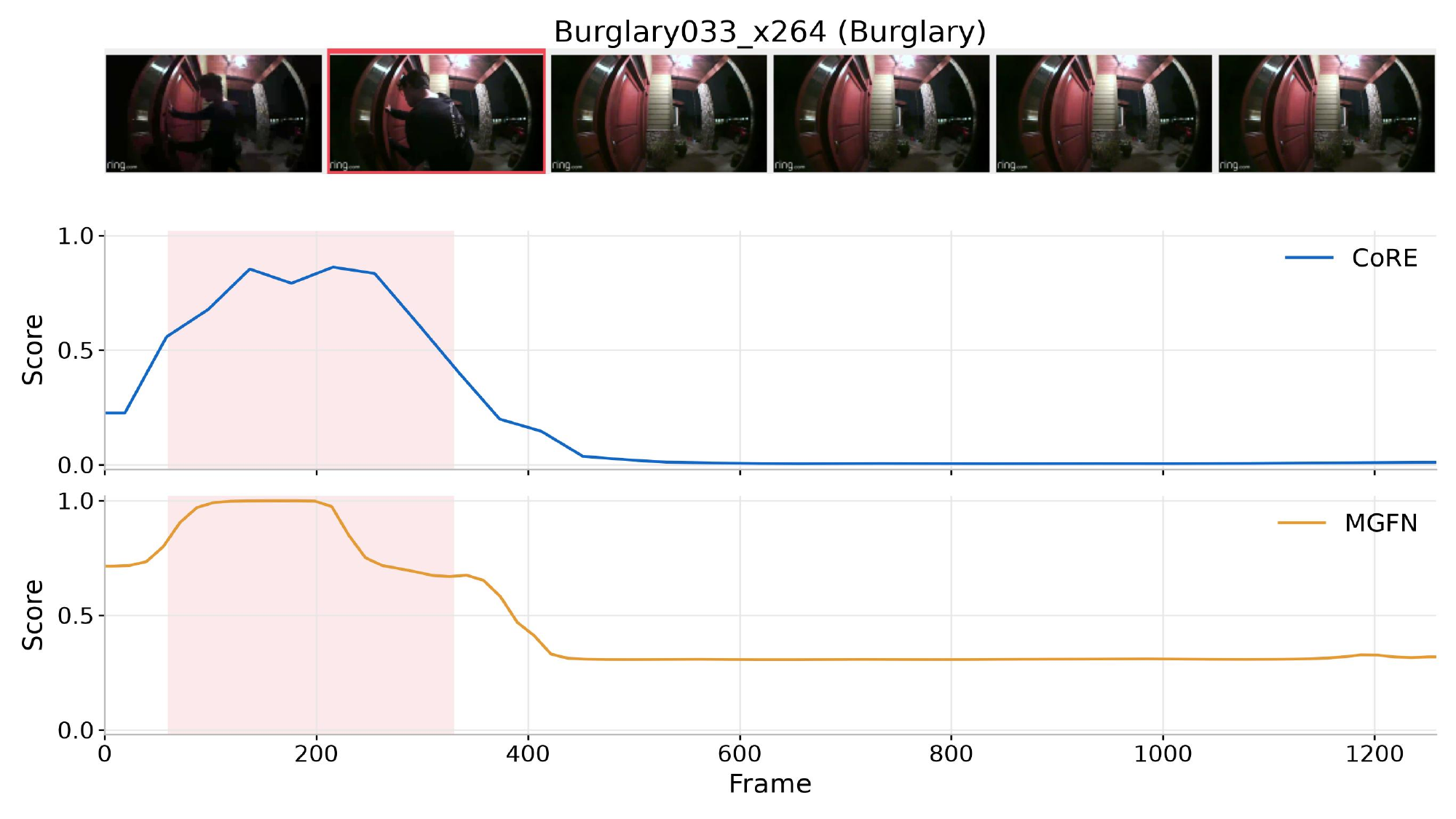}
    \hfill
    \includegraphics[width=0.44\textwidth]{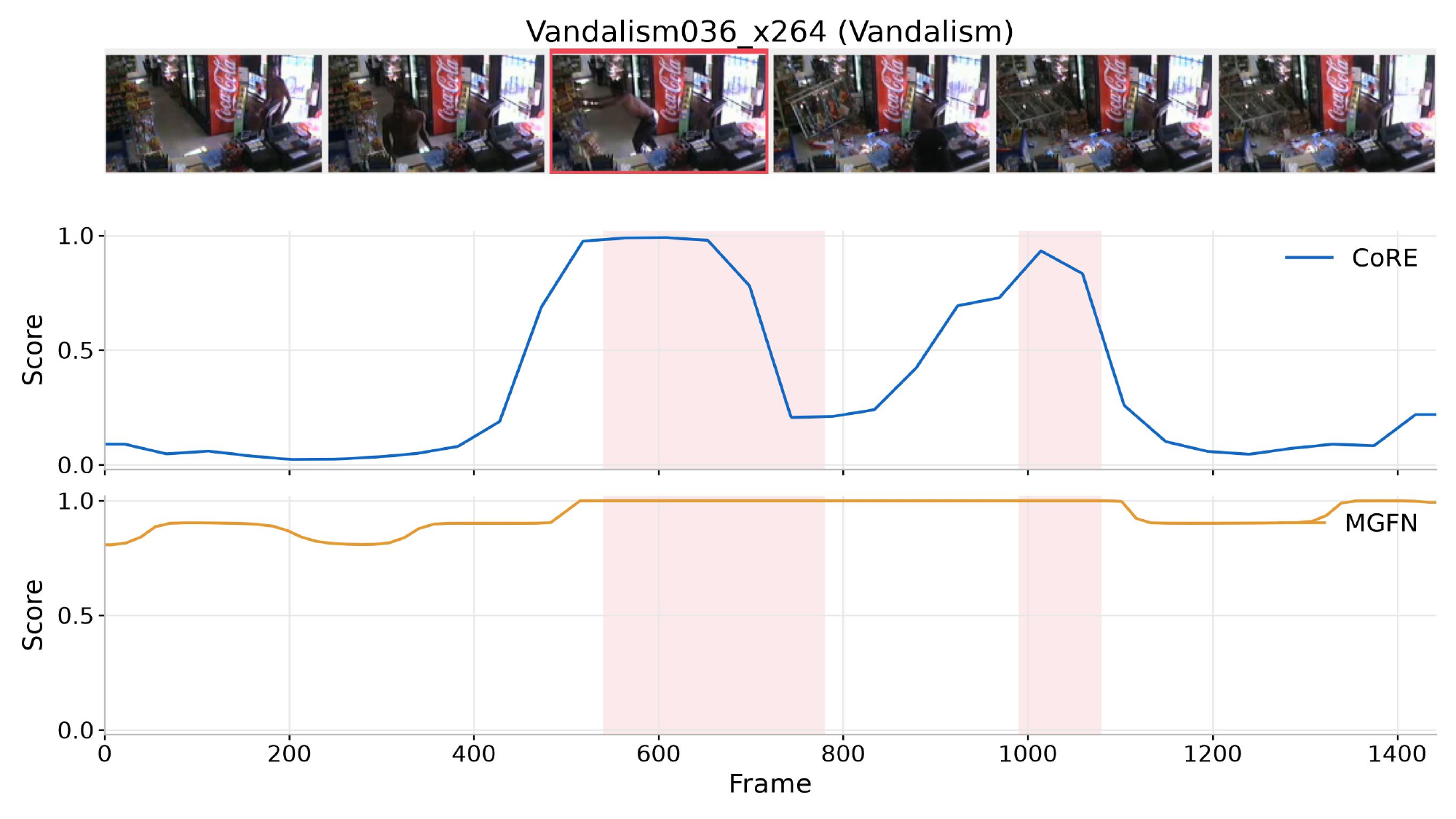}

    \caption{
    \textbf{Qualitative temporal localization on UCF-Crime.}
    Each example shows video frames, the frame-level anomaly interval
    (pink), and the temporal scores produced by CoRE and MGFN\cite{chen2022mgfnmagnitudecontrastiveglanceandfocusnetwork}.
    Red borders identify displayed frames within the annotated interval.
    }
    \label{fig:ucf_qualitative}
\end{figure*}
\noindent\textbf{Temporal localization on DoTA.}
Table~\ref{tab:dota_main_comparison} evaluates CoRE against temporal
annotations unseen during training. CoRE achieves the best
result on every metric under both feature banks. With
ResNet-50, frame AUC improves from $0.642$ to $0.735$, F1@0.5 from
$0.237$ to $0.374$, and best tIoU from $0.370$ to $0.440$. With
CLIP ViT-B/32, CoRE reaches $0.744$ AUC, $0.514$ AP, $0.364$
F1@0.5, and $0.429$ tIoU. Consistent gains across representations indicate that the
improvement is not tied to a particular feature bank.
DoTA's intervals are independent of the prediction effects used for
training, providing external validation that learned support
corresponds to meaningful event timing.

\begin{figure}[!b]
    \centering
    \includegraphics[width=\linewidth]{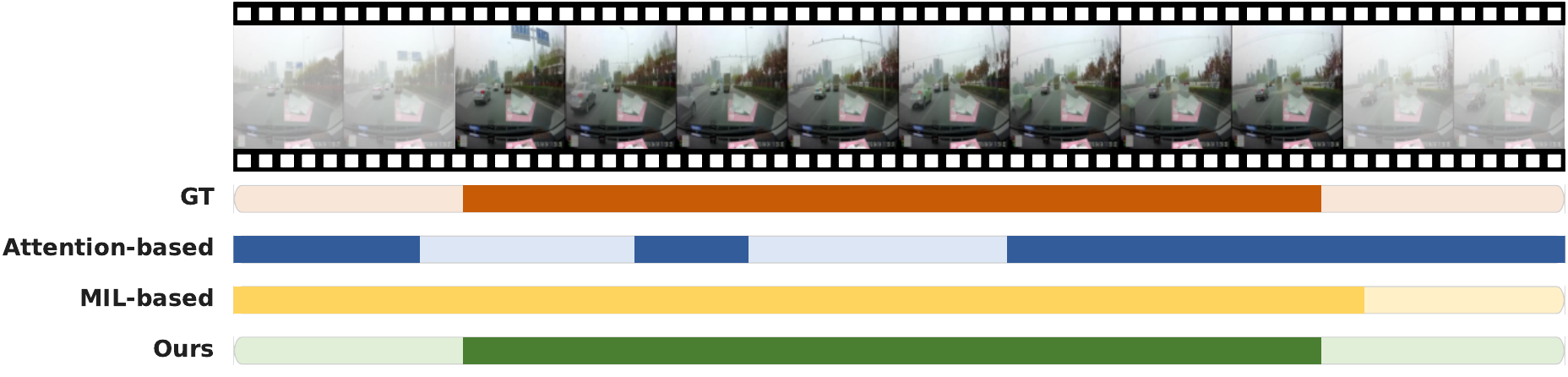}
    \vspace{0.35em}
    \includegraphics[width=\linewidth]{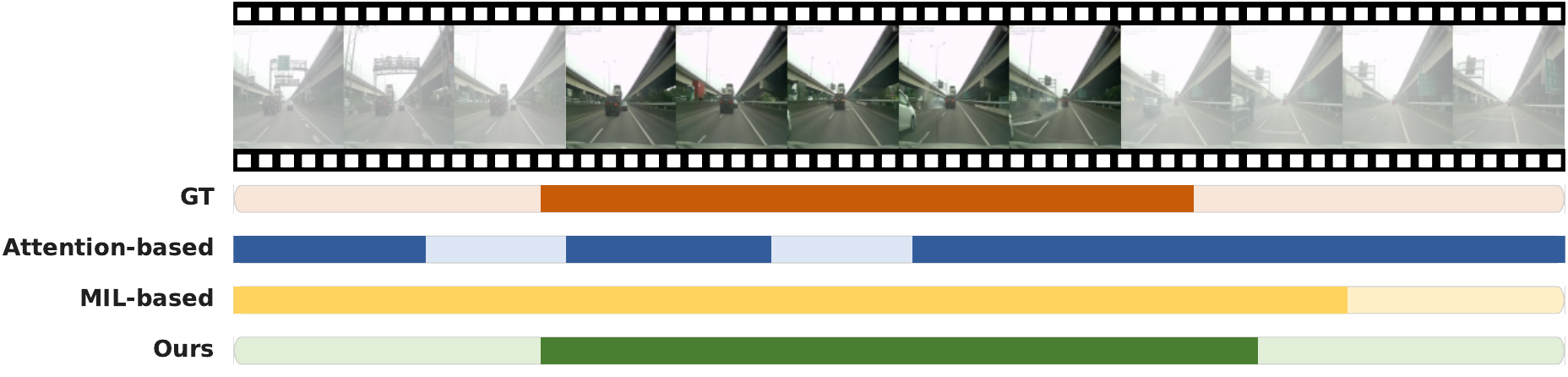}
    \vspace{0.35em}
    \includegraphics[width=\linewidth]{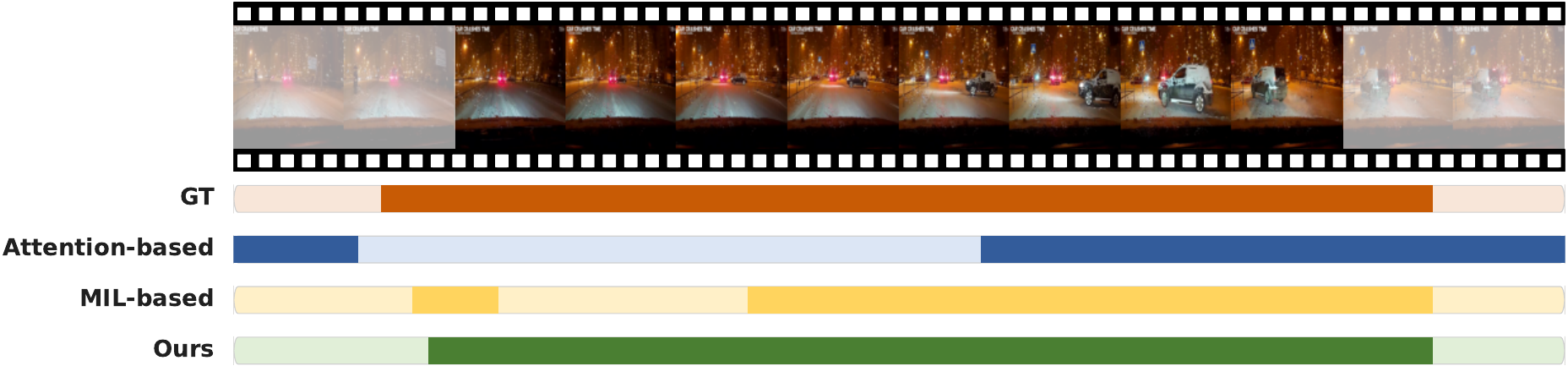}
    \caption{
    \textbf{Qualitative temporal localization on DoTA.}
    Ground-truth intervals are compared with attention-based OE-CTST~\cite{majhi2024oe}, MIL-based PE-MIL~\cite{chen2024prompt}, and CoRE on the same videos.
}
    \label{fig:dota_qual}
\end{figure}

\begin{figure}[!b]
    \centering
    \includegraphics[width=\columnwidth]{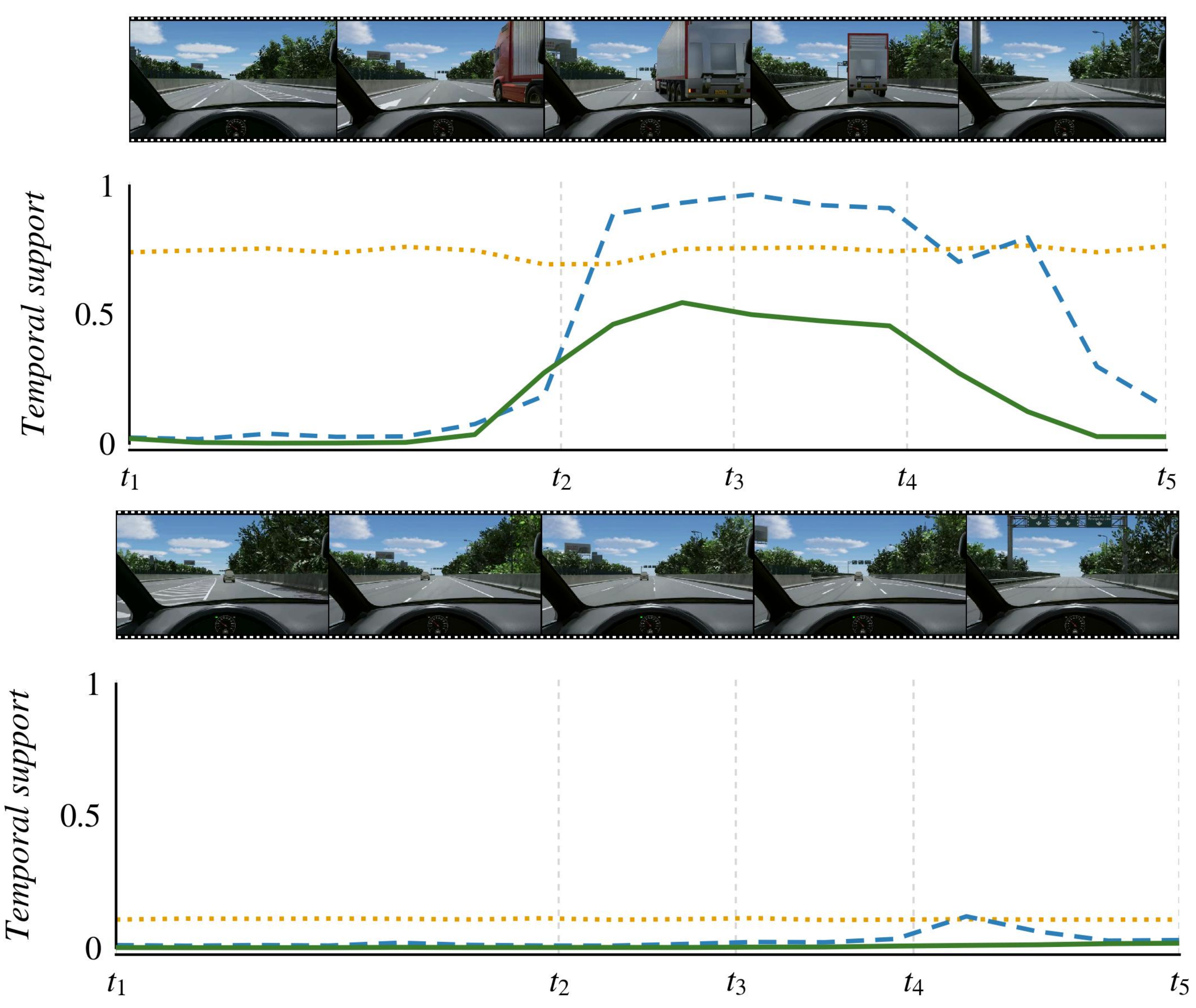}
    \caption{\textbf{Temporal support on contrasting RISEE clips.}
    $t_1$--$t_5$ mark the displayed frames.
    \mbox{\legendline{attnblue,dashed} Attn.\ MIL},
    \mbox{\legendline{topkgold,densely dotted} soft top-$k$},
    \mbox{\legendline{coregreen,solid} CoRE}.}
    \label{fig:risee_temporal_support}
\end{figure}

\noindent\textbf{Generalization to Non-Driving Task.}
Table~\ref{tab:ucf_crime} evaluates CoRE on non-driving anomaly
localization. CoRE achieves $85.68\%$ frame AUC, improving over
RTFM~\cite{tian2021weakly} and
MGFN~\cite{chen2022mgfnmagnitudecontrastiveglanceandfocusnetwork} by
$1.38$ and $2.89$ points, respectively. It also slightly exceeds
NG-MIL~\cite{10030221} ($85.63\%$) and is within $0.69$ points of
DSS~\cite{xu2025discriminative} ($86.37\%$). Methods using additional
semantic or visual information~\cite{chen2024prompt,majhi2025just,
wang2025learning,wang2026weakly} achieve between $86.83\%$ and
$91.05\%$ AUC. Overall, the results extend CoRE from subjective
driving-risk prediction to anomaly localization in long surveillance
videos with diverse scenes and anomaly categories.
\begin{table}[t]
\centering
\caption{\textbf{Weakly supervised anomaly detection on UCF-Crime.}
Frame-level AUC (\%). Methods are grouped by training information.
$^\dagger$ denotes results obtained by running the official
implementation. Best and second best within each group are shown in
\textbf{bold} and \underline{underlined}.}
\label{tab:ucf_crime}
\scriptsize
\setlength{\tabcolsep}{3pt}
\begin{tabular}{llcc}
\toprule
Method & Venue & Feature & AUC (\%) $\uparrow$ \\
\midrule
\multicolumn{4}{l}{\textit{Single visual representation, video-level labels only}}\\
MIL-Rank~\cite{sultani2018real}
    & CVPR'18 & C3D RGB & 75.41 \\
MIST~\cite{feng2021mist}
    & CVPR'21 & I3D RGB & 82.30 \\
RTFM$^\dagger$~\cite{tian2021weakly}
    & ICCV'21 & I3D RGB & 84.30 \\
NG-MIL~\cite{10030221}
    & WACV'23 & I3D RGB & 85.63 \\
MGFN$^\dagger$~\cite{chen2022mgfnmagnitudecontrastiveglanceandfocusnetwork}
    & AAAI'23 & I3D RGB & 82.79 \\
DSS~\cite{xu2025discriminative}
    & WACV'25 & I3D RGB & \textbf{86.37} \\
\addlinespace[1pt]
\textbf{\method{} (ours)}
    & -- & I3D RGB & \underline{85.68} \\
\addlinespace

\multicolumn{4}{l}{\textit{Additional modality or semantic information during training}}\\
PE-MIL~\cite{chen2024prompt}
    & CVPR'24 & I3D+text & 86.83 \\
$\pi$-VAD~\cite{majhi2025just}
    & CVPR'25 & I3D$^{\S}$ & \underline{90.33} \\
LEC-VAD~\cite{wang2025learning}
    & ICML'25 & I3D+text & 88.21 \\
LAS-VAD~\cite{wang2026weakly}
    & CVPR'26 & I3D+text & \textbf{91.05} \\
\bottomrule
\end{tabular}

\vspace{1mm}
\parbox{\columnwidth}{\scriptsize
$^{\S}$Uses pose, depth, panoptic, optical-flow, and VLM features
during training.
}
\end{table}

\subsection{Ablation and Analysis}
\label{sec:ablation}

\noindent\textbf{Component ablation.}
We ablate CoRE on DoTA using fixed CLIP ViT-B/32 features, data split,
model capacity, and optimization
(Table~\ref{tab:dota_ablation}). Removing prediction-effect
supervision causes the largest overall degradation, reducing AUC from
$0.744$ to $0.419$ and F1@0.5 from $0.364$ to $0.139$.
Removing graded effect targets likewise substantially degrades
performance, while removing multi-scale candidates nearly eliminates
event localization ($0.040$ F1@0.5 and $0.098$ tIoU). Removing the bag objective slightly improves frame AUC/AP but lowers
F1@0.5 and tIoU, indicating that it primarily benefits event-level
localization rather than framewise ranking. Overall, the full model
provides the strongest localization-sensitive performance.

\begin{table}[!tb]
\centering
\caption{\textbf{Component ablation on DoTA.} Each row removes one component of
\method{}; the full configuration is shown in the last row. Best per column in
\textbf{bold}.}
\label{tab:dota_ablation}
\scriptsize
\setlength{\tabcolsep}{5.0pt}
\renewcommand{\arraystretch}{1.08}
\resizebox{\columnwidth}{!}{%
\begin{tabular}{lcccc}
\toprule
Configuration
& AUC $\uparrow$
& AP $\uparrow$
& F1@0.5 $\uparrow$
& Best tIoU $\uparrow$ \\
\midrule
w/o effect supervision       & 0.419 & 0.264 & 0.139 & 0.329 \\
w/o bag objective            & \textbf{0.754} & \textbf{0.518} & 0.309 & 0.335 \\
w/o soft targets             & 0.453 & 0.283 & 0.133 & 0.326 \\
w/o multi-scale candidates   & 0.504 & 0.314 & 0.040 & 0.098 \\
\midrule
\textbf{Full \method{}}      & 0.744 & 0.514 & \textbf{0.364} & \textbf{0.429} \\
\bottomrule
\end{tabular}%
}
\end{table}

\noindent\textbf{Entity support.}
On RISEE, CoRE achieves a $0.243$ selected entity-effect drop and
$0.117$ gain over random; detailed design comparisons are provided
in the supplementary material.



\noindent\textbf{Qualitative analysis.}
Figure~\ref{fig:dota_qual} shows that \method{} concentrates support around
DoTA ground-truth events, while competing methods produce fragmented
responses or remain active outside the event interval. On UCF-Crime,
Figure~\ref{fig:ucf_qualitative} shows that \method{} produces event-centered
responses despite the different surveillance setting, while MGFN is more
diffuse. For RISEE, frame-level human annotations are unavailable, so
Figure~\ref{fig:risee_temporal_support} illustrates inferred temporal support
under clip-level supervision. Both \method{} and Attention MIL increase
support as the high-risk interaction develops and maintain low support
throughout the low-risk sequence. Soft top-$k$ distinguishes the clips
through different support levels but exhibits limited temporal variation.
These observations complement the intervention-based analysis in
Table~\ref{tab:risee_temporal}.

\section{Conclusion}
We presented \textbf{CoRE}, a weakly supervised framework that learns
fine-grained support from prediction changes under structured
interventions. A student distills these effects to predict temporal
and entity support without interventions at inference. RISEE
demonstrates learning from clip-level perceived-risk judgments,
DoTA validates temporal support against independent event
annotations, and UCF-Crime extends the framework beyond driving.
CoRE turns coarse video supervision into predictions with
supporting evidence.

\section*{LLM Usage Statement}
LLMs were used only to polish author-written text for grammar and clarity, and to help 
refine implementation code. No LLMs or AI tools were used to generate research ideas, 
conduct experiments, create figures or datasets, or review the manuscript.

{
    \small
    \bibliographystyle{unsrtnat}
    \bibliography{main}
}

\clearpage
\setcounter{page}{1}
\maketitlesupplementary
\appendix
\suppressfloats[t]

\noindent\textbf{Overview.}
This supplement presents additional experiments on entity-support
design (Sec.~A), masking methods and coarse predictors (Sec.~B),
and direct attention from the same teacher (Sec.~C).
We further report runtime (Sec.~D), temporal localization against
expert annotations on RISEE (Sec.~E), and entity localization
on DRAMA (Sec.~F). The accompanying codebase provides the
implementation, and configurations.

\section{Entity-Support Design}
\label{sec:supp_entity}

Table~\ref{tab:risee_object} compares entity-support designs using the same candidate tracks. Hard targets use the highest-effect track for training, while soft targets retain scores for all tracks. \method{} achieves the highest Sel. Drop ($0.243$) and Gain/Rand. ($0.117$), and the lowest Regret@1 ($0.087$). Hard targets lead in Effect $\rho$ ($0.425$) and NDCG@3 ($0.747$). Adding interaction and temporal conditioning gives similar selection results.
\begin{figure*}[t]
    \centering
    \includegraphics[width=0.9\textwidth]{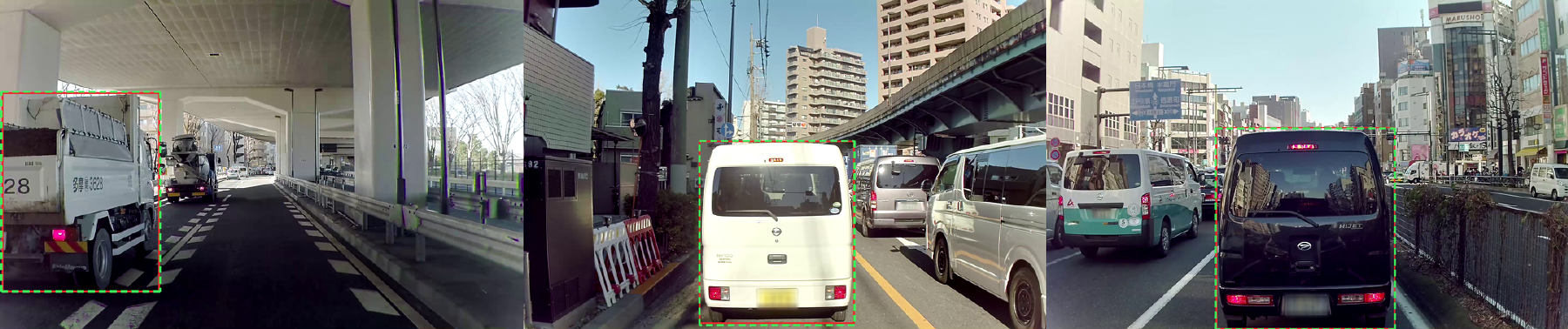}
    \caption{\textbf{Entity localization on DRAMA.}
    Red boxes show CoRE predictions and green boxes indicate
    ground-truth important objects.}
    \label{fig:drama_qual}
\end{figure*}
\begin{table}[t]
\centering
\caption{\textbf{Entity-support design on RISEE.} All methods use the same candidate tracks. Best in \textbf{bold}, second best \underline{underlined}.}
\label{tab:risee_object}
\scriptsize
\setlength{\tabcolsep}{3pt}
\resizebox{\columnwidth}{!}{%
\begin{tabular}{lccccc}
\toprule
Method & Sel. Drop $\uparrow$ & Gain/Rand. $\uparrow$ & Effect $\rho$ $\uparrow$ & NDCG@3 $\uparrow$ & Regret@1 $\downarrow$ \\
\midrule
Hard top-track target & 0.220 & 0.094 & \textbf{0.425} & \textbf{0.747} & 0.110 \\
Soft effect distribution & 0.237 & 0.111 & 0.381 & 0.739 & 0.093 \\
\method{} + interaction & 0.232 & 0.106 & 0.358 & 0.717 & 0.098 \\
\method{} + interaction + temporal & \underline{0.242} & \underline{0.116} & 0.348 & 0.727 & \underline{0.088} \\
\midrule
\textbf{\method{} (ours)} & \textbf{0.243} & \textbf{0.117} & \underline{0.388} & \underline{0.746} & \textbf{0.087} \\
\bottomrule
\end{tabular}%
}
\end{table}

\section{Masking and Coarse Predictors}
\label{sec:supp_masking}

\noindent\textbf{Masking methods.}
Table~\ref{tab:cross_operator} compares local-mean replacement and blur for target construction and evaluation. Changing the training operator has little effect under a fixed evaluation operator. For temporal support, Sel. Drop changes from $0.0826$ to $0.0810$ under mean-replacement evaluation and from $0.0706$ to $0.0705$ under blur evaluation. The entity results show the same pattern, with positive Gain/Rand. in every setting. This comparison reports prediction changes on the normalized risk scale.

\begin{table}[t]
\centering
\caption{\textbf{Masking methods on RISEE.} Train $\rightarrow$ Eval denotes target construction and evaluation. Mean denotes local-mean replacement.}
\label{tab:cross_operator}
\scriptsize
\setlength{\tabcolsep}{5pt}
\resizebox{\columnwidth}{!}{%
\begin{tabular}{llcc}
\toprule
Support & Train $\rightarrow$ Eval & Sel. Drop $\uparrow$ & Gain/Rand. $\uparrow$ \\
\midrule
Temporal & Mean $\rightarrow$ Mean & 0.0826 & 0.0558 \\
Temporal & Blur $\rightarrow$ Mean & 0.0810 & 0.0542 \\
Temporal & Mean $\rightarrow$ Blur & 0.0706 & 0.0418 \\
Temporal & Blur $\rightarrow$ Blur & 0.0705 & 0.0417 \\
\midrule
Entity & Mean $\rightarrow$ Mean & 0.0615 & 0.0284 \\
Entity & Blur $\rightarrow$ Mean & 0.0588 & 0.0257 \\
Entity & Mean $\rightarrow$ Blur & 0.0598 & 0.0282 \\
Entity & Blur $\rightarrow$ Blur & 0.0579 & 0.0263 \\
\bottomrule
\end{tabular}%
}
\end{table}

\noindent\textbf{Agreement between masking methods.}
Local-mean replacement and blur produce similar rankings of candidate effects (Table~\ref{tab:operator_agreement}). Their Spearman correlations are $0.7344$ for temporal windows and $0.7356$ for entity tracks. The two operators select the same highest-effect candidate in $60.89\%$ of temporal cases and $83.33\%$ of entity cases.

\begin{table}[t]
\centering
\caption{\textbf{Agreement between masking methods on RISEE.}}
\label{tab:operator_agreement}
\scriptsize
\setlength{\tabcolsep}{6pt}
\begin{tabular}{lcc}
\toprule
Support & Spearman $\rho$ $\uparrow$ & Agree@1 $\uparrow$ \\
\midrule
Temporal & 0.7344 & 0.6089 \\
Entity & 0.7356 & 0.8333 \\
\bottomrule
\end{tabular}
\end{table}

\noindent\textbf{Coarse predictors.}
Table~\ref{tab:teacher_robustness} compares support students trained from three coarse predictors using fixed candidates and student design. All three give positive Gain/Rand. The CoRE coarse predictor has the highest Sel. Drop ($2.5256$) and Effect $\rho$ ($0.7956$), while the Attention MIL predictor gives the highest Gain/Rand. ($1.7356$).

\begin{table}[t]
\centering
\caption{\textbf{Coarse-predictor comparison on RISEE.} Candidate sets and student design are fixed.}
\label{tab:teacher_robustness}
\scriptsize
\setlength{\tabcolsep}{5pt}
\resizebox{\columnwidth}{!}{%
\begin{tabular}{lccc}
\toprule
Teacher & Sel. Drop $\uparrow$ & Gain/Rand. $\uparrow$ & Effect $\rho$ $\uparrow$ \\
\midrule
Mean Pool & 0.8519 & 0.5575 & 0.7417 \\
Attention MIL & 2.4585 & \textbf{1.7356} & 0.6028 \\
CoRE coarse predictor & \textbf{2.5256} & 1.6813 & \textbf{0.7956} \\
\bottomrule
\end{tabular}%
}
\end{table}

\section{CoRE Versus Attention}
\label{sec:supp_attention}

Table~\ref{tab:attention_vs_core} compares direct attention with CoRE support learned from the same Attention MIL teacher~\cite{ilse2018attention}. CoRE improves Sel. Drop from $2.3295$ to $2.4585$ and Effect $\rho$ from $0.5683$ to $0.6028$. It also improves Gain/Rand., NDCG@3, and Top Perc. in this matched-teacher comparison.

\begin{table}[t]
\centering
\caption{\textbf{CoRE versus attention on RISEE.} Both outputs use the same Attention MIL teacher.}
\label{tab:attention_vs_core}
\scriptsize
\setlength{\tabcolsep}{4.5pt}
\resizebox{\columnwidth}{!}{%
\begin{tabular}{lccccc}
\toprule
Support & Sel. Drop $\uparrow$ & Gain/Rand. $\uparrow$ & Effect $\rho$ $\uparrow$ & NDCG@3 $\uparrow$ & Top Perc. $\uparrow$ \\
\midrule
Attention MIL attention & 2.3295 & 1.6066 & 0.5683 & 0.8211 & 0.8823 \\
\textbf{CoRE support} & \textbf{2.4585} & \textbf{1.7356} & \textbf{0.6028} & \textbf{0.8549} & \textbf{0.9082} \\
\bottomrule
\end{tabular}%
}
\end{table}

\section{Runtime}
\label{sec:supp_runtime}

Table~\ref{tab:runtime} reports runtime on an NVIDIA H100 NVL. The coarse predictor and temporal student heads each take about $2.07$\,ms, and the entity student head takes $0.55$\,ms. ResNet-50 feature extraction for 16 frames takes $10.87$\,ms. Candidate-wise intervention scoring takes $14.22$\,ms during target construction. CoRE predicts support directly at inference.

\begin{table}[t]
\centering
\caption{\textbf{Runtime on NVIDIA H100 NVL.} Times are in ms and peak GPU memory in MB.}
\label{tab:runtime}
\scriptsize
\setlength{\tabcolsep}{4pt}
\resizebox{\columnwidth}{!}{%
\begin{tabular}{lccc}
\toprule
Component & Mean (ms) $\downarrow$ & p95 (ms) $\downarrow$ & Peak GPU (MB) $\downarrow$ \\
\midrule
Coarse teacher head & 2.0679 & 2.0836 & 82.87 \\
Temporal student head & 2.0676 & 2.0833 & 82.87 \\
Entity student head & 0.5510 & 0.5663 & 175.92 \\
ResNet-50, 16 frames & 10.8651 & 10.8832 & 260.82 \\
Offline temporal target construction & 14.2064 & 14.2444 & 88.10 \\
Candidate-wise intervention scoring & 14.2172 & 14.2654 & 88.10 \\
\bottomrule
\end{tabular}%
}
\end{table}

\section{Expert Temporal Evaluation}
\label{sec:supp_human}

Three experts independently annotated risk presence and its start and end times in 30 RISEE videos. For the 27 videos with a consensus interval, we use the median annotated start and end times. The remaining three videos have no consensus interval. We compare out-of-fold predictions from CoRE, Attention MIL, and Soft top-$k$ MIL using the same temporal candidates. Expert annotations are used only for evaluation.

\begin{figure*}[t]
\centering
\includegraphics[width=0.85\textwidth]{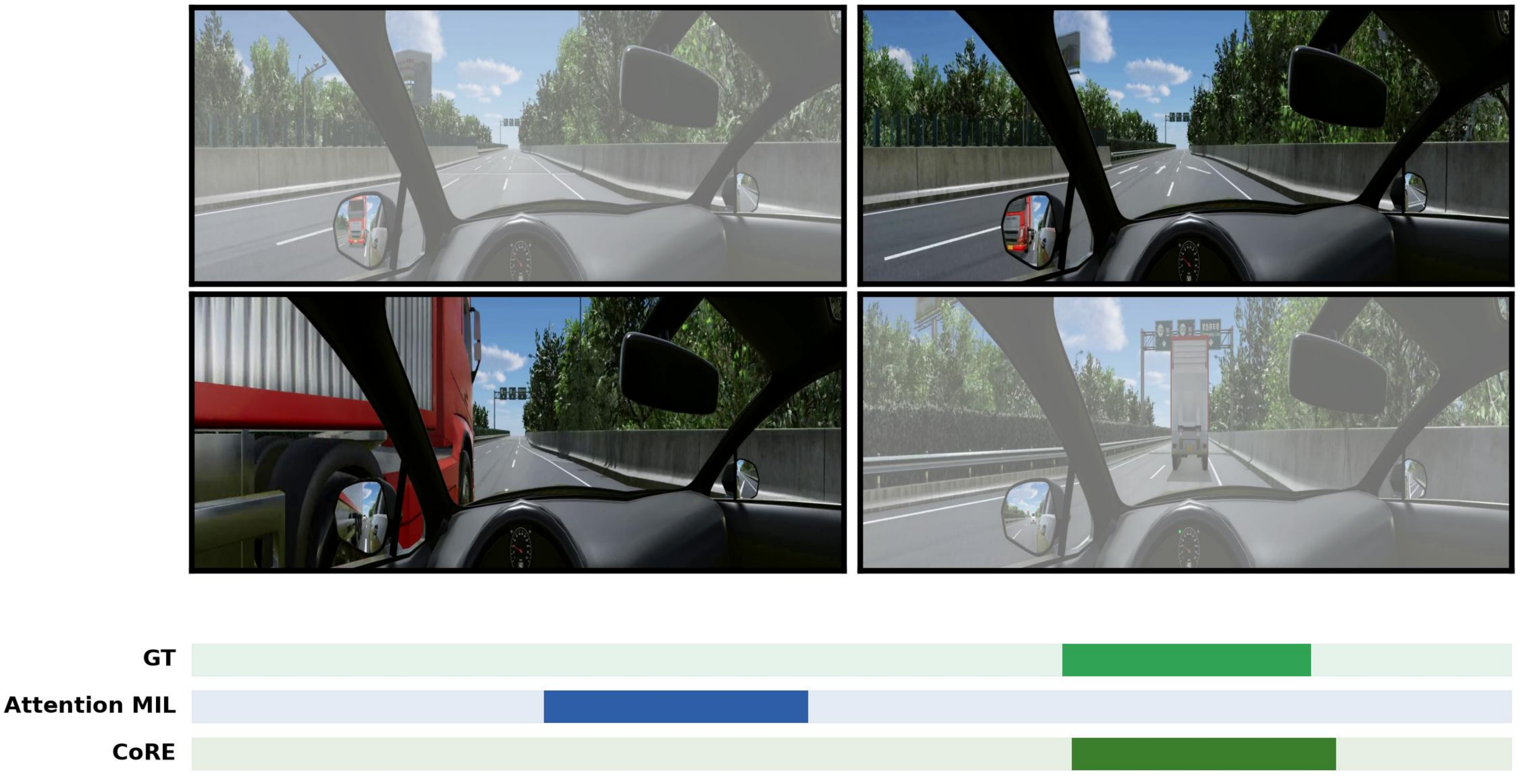}
\vspace{0.6em}
\includegraphics[width=0.85\textwidth]{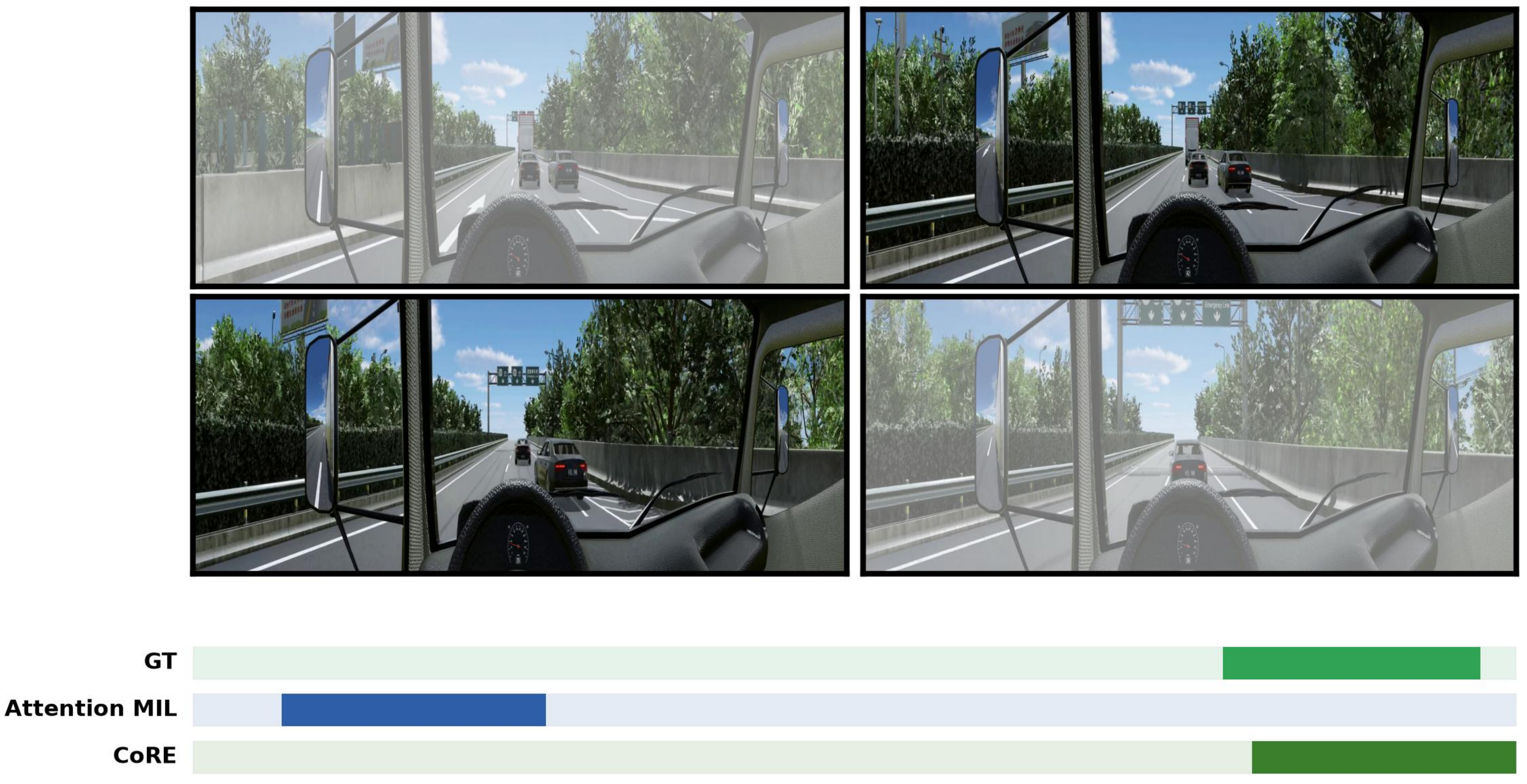}
\vspace{0.6em}
\includegraphics[width=0.85\textwidth]{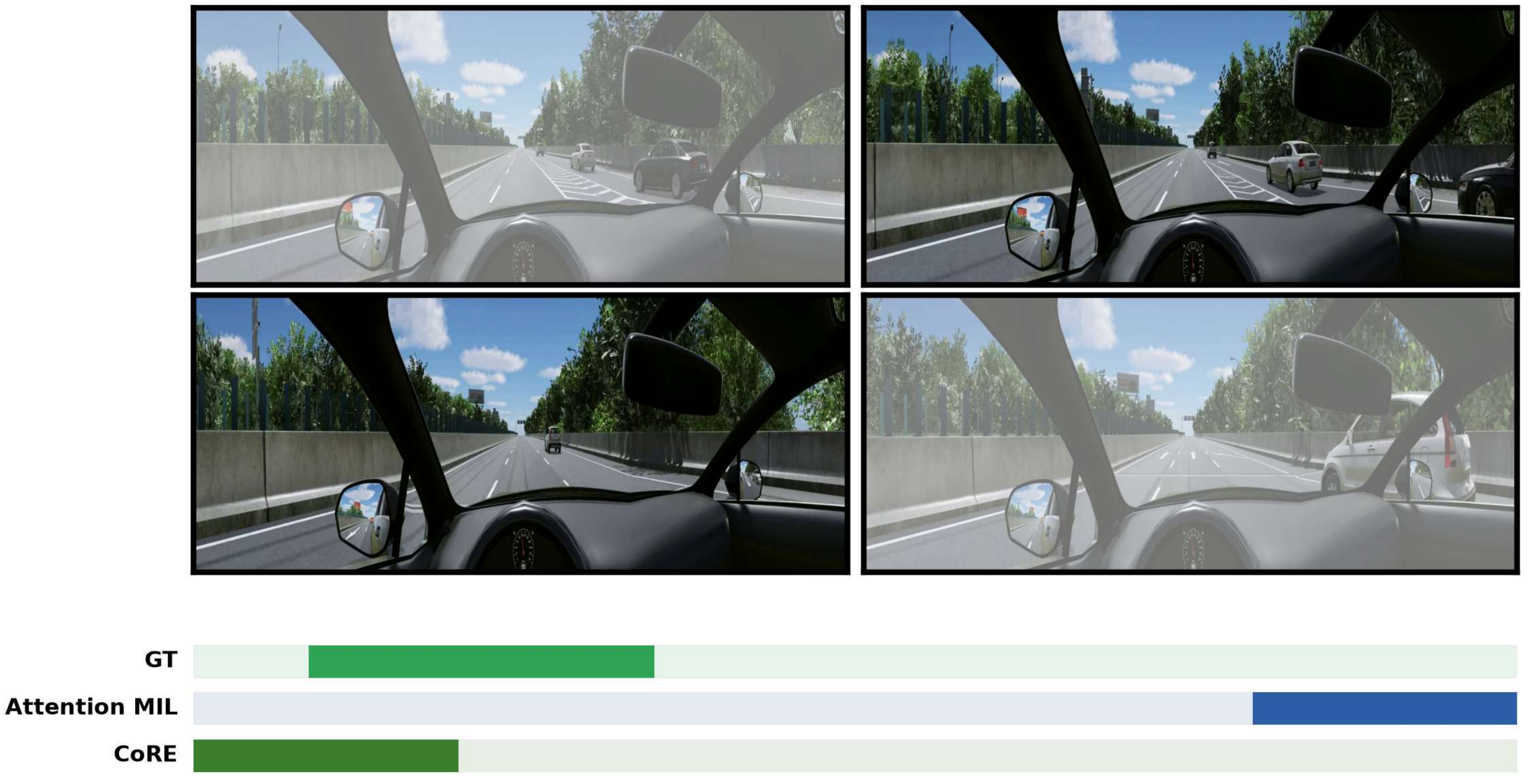}
\caption{\textbf{Temporal localization on RISEE.} Expert-consensus intervals and predictions from Attention MIL and CoRE. The third example shows partial overlap.}
\label{fig:risee_human_qual}
\end{figure*}

Table~\ref{tab:risee_human_temporal} compares the temporal predictions with the expert intervals. CoRE reaches a mean temporal IoU (tIoU) of $0.604$, compared with $0.204$ for Soft top-$k$ MIL and $0.118$ for Attention MIL. R@.3 and R@.5 measure recall at tIoU thresholds of $0.3$ and $0.5$. CoRE reaches $0.704$ on R@.5, with mean onset and end errors of $1.32$\,s and $1.18$\,s. Figure~\ref{fig:risee_human_qual} shows two closely aligned examples and one with partial overlap.

\begin{table}[t]
\centering
\small
\setlength{\tabcolsep}{1.8pt}
\renewcommand{\arraystretch}{1.05}
\caption{\textbf{Temporal localization on RISEE.} Evaluation against expert-consensus intervals. Boundary errors are in seconds.}
\label{tab:risee_human_temporal}
\resizebox{\columnwidth}{!}{%
\begin{tabular}{lcccccccc}
\toprule
Method
& \begin{tabular}[c]{@{}c@{}}Frame\\AUC $\uparrow$\end{tabular}
& \begin{tabular}[c]{@{}c@{}}Frame\\AP $\uparrow$\end{tabular}
& \begin{tabular}[c]{@{}c@{}}Macro\\AUC $\uparrow$\end{tabular}
& tIoU $\uparrow$
& R@.3 $\uparrow$
& R@.5 $\uparrow$
& \begin{tabular}[c]{@{}c@{}}Onset\\Err. $\downarrow$\end{tabular}
& \begin{tabular}[c]{@{}c@{}}End\\Err. $\downarrow$\end{tabular} \\
\midrule
Attention MIL~\cite{ilse2018attention}
& .448 & .196 & .438 & .118 & .185 & .074 & 5.87 & 5.94 \\
Soft top-$k$ MIL~\cite{sultani2018real}
& .596 & .297 & .549 & .204 & .370 & .148 & 4.42 & 4.69 \\
\textbf{CoRE (ours)}
& \textbf{.700} & \textbf{.451} & \textbf{.900} & \textbf{.604}
& \textbf{.926} & \textbf{.704} & \textbf{1.32} & \textbf{1.18} \\
\bottomrule
\end{tabular}%
}
\end{table}

\section{Entity Support on DRAMA}
\label{sec:supp_drama}

\noindent\textbf{Dataset and setup.}
DRAMA~\cite{malla2023drama} contains 17,785 driving
scenarios with important-object annotations and captions.
We use 12,442 videos for training, 2,666 for validation,
and 2,669 for testing. All methods share frozen CLIP
ViT-B/32 features, tracked entity candidates, and the same
temporal architecture. We adapt Attention
MIL~\cite{ilse2018attention} and Soft top-$k$
MIL~\cite{sultani2018real} to predict entity support
using video-level risk labels. CoRE learns entity support
from the prediction effects of intervening on individual
entity candidates.

\noindent\textbf{Entity localization.}
Table~\ref{tab:drama_entity} evaluates entity support
against annotated important-object boxes. \method{} achieves
67.49\% mIoU and 74.79\% Acc@0.5, improving over
Attention MIL by 6.09 and 7.39 points, respectively.
\method{} also outperforms LCP on both metrics.

\noindent\textbf{Qualitative results.}
Figure~\ref{fig:drama_qual} shows entity localization on
DRAMA. CoRE identifies the relevant vehicles across different
driving scenes, with predicted boxes closely matching the
ground-truth annotations.

\begin{table}[t]
\centering
\caption{\textbf{Entity localization on DRAMA.}
Mean IoU and Acc@0.5 (\%).}
\label{tab:drama_entity}
\small
\setlength{\tabcolsep}{6pt}
\begin{tabular}{lcc}
\toprule
Method & mIoU $\uparrow$ & Acc@0.5 $\uparrow$ \\
\midrule
\multicolumn{3}{l}{\textit{Object annotations and captions}} \\
ICL~\cite{malla2023drama} & 53.30 & 59.30 \\
CLP~\cite{malla2023drama} & 52.00 & 58.10 \\
LCP~\cite{malla2023drama} & 61.40 & 68.40 \\
\midrule
\multicolumn{3}{l}{\textit{Video-level supervision}} \\
Attention MIL~\cite{ilse2018attention}
& 61.40 & 67.40 \\
Soft top-$k$ MIL~\cite{sultani2018real}
& 60.63 & 66.46 \\
\textbf{\method{} (ours)}
& \textbf{67.49} & \textbf{74.79} \\
\bottomrule
\end{tabular}
\end{table}


\end{document}